%% file: neurips_2025.tex
\documentclass{article}

\PassOptionsToPackage{numbers, compress,sort}{natbib}

\usepackage[final]{neurips_2025}

\usepackage{enumitem}
\usepackage[utf8]{inputenc} 
\usepackage[T1]{fontenc}    
\usepackage{hyperref}       
\usepackage{url}            
\usepackage{booktabs}       
\usepackage{amsfonts}       
\usepackage{amsmath}
\usepackage{multirow}
\usepackage{graphicx}
\usepackage{nicefrac}       
\usepackage{microtype}      
\usepackage{xcolor}         

\usepackage{algorithm}
\usepackage{listings}
\usepackage{algpseudocode}
\usepackage{arydshln}
\usepackage{wrapfig,booktabs}
\usepackage{pifont}
\usepackage{subcaption}
\newcommand{\eg}{\textit{e.g.}}
\newcommand{\ie}{\textit{i.e.}}
\hypersetup{colorlinks,citecolor={blue}}

\definecolor{Gray}{gray}{0.9}

\usepackage{xspace}

\title{Training-free Detection of AI-generated images via Cropping Robustness}

\author{%
Sungik Choi$^{1}$  \quad Hankook Lee$^{2}$ \quad Moontae Lee$^{1,3}$ \\
$^1$LG AI Research \quad $^{2}$SungkyunKwan University \quad  $^3$University of Illinois Chicago \\
\texttt{sungik.choi@lgresearch.ai}\\
}

\def\rvx{{\mathbf{x}}}

\newcommand{\basename}
{$\text{HFwav}$\xspace}
\newcommand{\methodname}{$\text{WaRPAD}$\xspace}

\begin{document}

\maketitle

\begin{abstract}
\input{sec/0_abstract}
\end{abstract}

\input{sec/1_introduction}

\input{sec/2_preliminary}

\input{sec/3_method}

\input{sec/4_experiment}

\input{sec/6_conclusion}
\ack{This work is fully supported by LG AI Research.}
\newpage
\bibliography{main}
\bibliographystyle{unsrtnat}

\newpage
\input{sec/X_checklist}

\newpage
\appendix
\input{sec/Y_Appendix}

\end{document}

%% file: sec/0_abstract.tex
AI-generated image detection has become crucial with the rapid advancement of vision-generative models. Instead of training detectors tailored to specific datasets, we study a training-free approach leveraging self-supervised models without requiring prior data knowledge. These models, pre-trained with augmentations like \texttt{RandomResizedCrop}, learn to produce consistent representations across varying resolutions. Motivated by this, we propose \textbf{WaRPAD}, a training-free AI-generated image detection algorithm based on self-supervised models. Since neighborhood pixel differences in images are highly sensitive to resizing operations, WaRPAD first defines a base score function that quantifies the sensitivity of image embeddings to perturbations along high-frequency directions extracted via Haar wavelet decomposition. To simulate robustness against cropping augmentation, we rescale each image to a multiple of the model’s input size, divide it into smaller patches, and compute the base score for each patch. The final detection score is then obtained by averaging the scores across all patches. We validate WaRPAD on real datasets of diverse resolutions and domains, and images generated by 23 different generative models. Our method consistently achieves competitive performance and demonstrates strong robustness to test-time corruptions. Furthermore, as invariance to \texttt{RandomResizedCrop} is a common training scheme across self-supervised models, we show that WaRPAD is applicable across self-supervised models.    

%% file: sec/1_introduction.tex
\section{Introduction}

AI-generated image detection aims to design reliable metrics that can distinguish between real and synthetically generated images. This task has become increasingly critical with the advent of highly capable text-to-image (T2I) generative models that can produce photorealistic outputs, which may be exploited for vicious purposes (\eg, fake news \cite{howcroft2025aicontent}, deepfakes \cite{murphykelly2025deepfakes}). Most existing detection approaches \cite{Coozolino24MDL,Yan24Combine} are trained to recognize specific real data distributions (\eg, ImageNet \cite{Deng09ImageNet}, LSUN \cite{Yu15LSUN}). However, the scope of real image distributions that current detection approaches can effectively cover remains extremely limited compared to the diversity of generated images. Furthermore, a training dataset for detection may contain artifacts (\eg, WebP compression in LSUN), which may lead the detector to overfit the artifacts and may fail to be robust in test-time corruptions \cite{rajan2025staypositivecaseignoringreal}. Consequently, there is a growing need for detection methods that can operate universally across diverse domains without relying on the constraints of predefined real image distributions.

In response to this growing demand, this paper focuses on \emph{training-free} detection methods where no prior knowledge of real image distributions is given. Prior training-free detection methods have design score functions based on representations extracted from large-scale pre-trained foundation models. One representative line of work leverages the representation of latent diffusion models (LDMs) (\eg, Stable Diffusion~\cite{Rombach22SD}, Midjourney~\cite{22Midjourney}). For instance, AEROBLADE \citep{ricker24aeroblade} detects LDM-generated images by measuring the autoencoder reconstruction loss, while Manifold Bias \cite{Brokman25Manifold} proposes a curvature-based metric in the latent space of the LDM. However, the performance of these approaches is often tied closely to the choices of the LDMs, and their generalizability to other generative models remains uncertain. 

Another stream of training-free research utilizes representations from self-supervised models, such as DINOv2 \cite{oquab24dinov2}. These models benefit from pre-training on a wide range of real-world images and are often applied as generalist models for various downstream tasks \cite{Liu24SegmentDINO,Darcet24Register}. Some prior works \cite{He24rigid,Tsai24MINDER} attempt to utilize self-supervised models on AI-generated image detection by introducing simple perturbations (\eg, Gaussian noise or Gaussian blurring) and measure the cosine similarity between the original and perturbed image embeddings. However, their performance often lags behind the diffusion-based training-free baselines (see Section~\ref{ssec:main_results}).

\textbf{Contribution.} In this study, we approach AI-generated image detection from a data augmentation perspective, utilizing foundation models that have been pre-trained on real images. Our primary motivation stems from the observation that these models are trained to produce consistent embedding representations between an original image and its randomly cropped and resized variants. We hypothesize that embeddings of AI-generated images exhibit lower robustness to such \texttt{RandomResizedCrop} (RRC) transformations than those of real images.

To investigate this hypothesis, we examine how RRC affects the high-frequency components of images, as obtained through wavelet decomposition. Notably, we observe that even when the cropped image closely matches the original in size, RRC introduces substantial variations in the difference in the neighborhood pixels. This indicates that RRC acts as an effective perturbation on the high-frequency components extracted via Haar wavelet decomposition. Given that foundation models are typically trained to be invariant to such perturbations on the real images, we propose a detection score function that quantifies embedding sensitivity to high-frequency distortions as a signal for distinguishing real and synthetic images.

\input{fig/fig_intro}

Furthermore, to simulate the effects of RRC in a more structured and deterministic manner, we resize each image to a multiple ($\times K^{2}$) of the default input resolution and partition it into $K^{2}$ patches of the default resolution. We then apply the proposed score function to each patch and aggregate the results. This patch-based perturbation consistently reveals a greater discrepancy in AI-generated images, demonstrating the effectiveness of our method. Figure~\ref{fig:main_picture} shows the computation of our unified method, \textbf{WaRPAD}: \textbf{Wa}velet, \textbf{R}esizing, and \textbf{P}atchifying for \textbf{A}I-generated image \textbf{D}etection.

We conduct extensive evaluations of \methodname across multiple AI-generated image detection benchmarks. These benchmarks span various generative model types, including LDMs, proprietary models (\eg, Firefly \cite{Adobe25Firefly}, Dall-E  \cite{ramesh2022dalle2}), and generative adversarial network (GAN) \cite{Goodfellow14GAN} architectures, as well as multiple image domains. Our method consistently outperforms other training-free baselines in all settings. Notably, we observe an improvement of \textbf{6.5 $\sim$ 24.7\% in AUROC} over prior methods based on the same DINOv2 model. Furthermore, we evaluate the robustness of our method against various image corruptions and show that it maintains competitive performance under such conditions, surpassing other detection methods.

In brief, our contributions are summarized as follows.

\begin{itemize}[leftmargin=15pt,topsep=0pt]
\setlength\itemsep{0em}
    \item We propose \methodname that applies a self-supervised foundation model's robustness on RRC augmentation for detecting AI-generated images (Section~\ref{sec:method}). 

    \item \methodname outperforms existing training-free AI-generated image detection methods in every benchmark consistently (Section~\ref{ssec:main_results}). Furthermore, \methodname is robust to test-time corruption of the examined images (Section~\ref{ssec:ablation_studies}).

    \item Our analysis suggests that a self-supervised model trained to be invariant under RRC can be applied for AI-generated image detection, supporting the generalizability of \methodname (Section~\ref{ssec:ablation_studies}).
\end{itemize}

%% file: fig/fig_intro.tex
\begin{figure}[t]
    \includegraphics[width=0.98\textwidth]{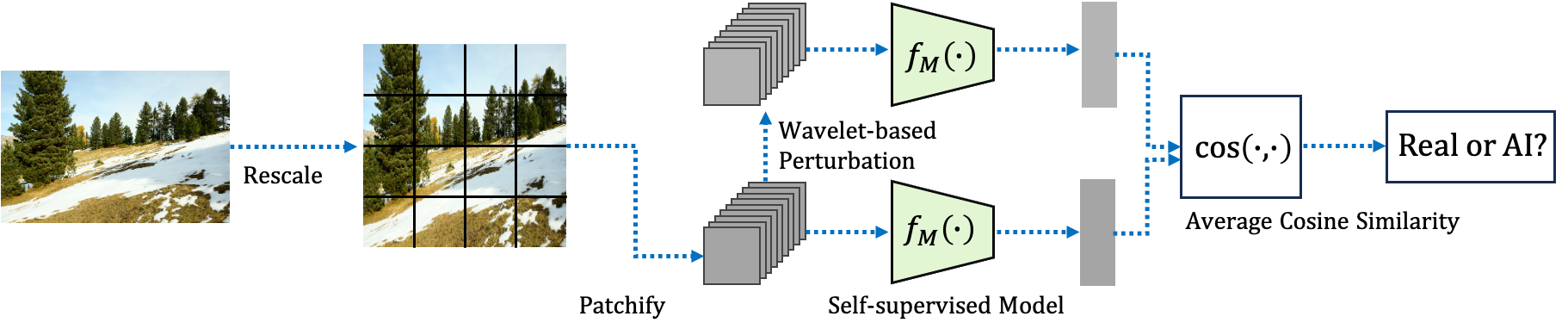}
    \caption{\textbf{Conceptual illustration of our method \methodname.} We first rescale and patchify the given image to the batch of patches. Then, we perturb the patches on the high-frequency direction of Haar wavelet decomposition. Our final score function is the averaged cosine similarity between the perturbed and non-perturbed patches' features through the self-supervised model.}
    \label{fig:main_picture}
\vspace{-0.1in}
\end{figure}

%% file: sec/2_preliminary.tex
\section{Preliminary}

First, we introduce the AI-generated image detection framework. Furthermore, we discuss the self-supervised models and their key data augmentation strategy. For the broader overview, we refer to \citet{deng2024survey} and \citet{uelwer2024survey} for the comprehensive survey.

\subsection{AI-generated Image Detection}

AI-generated image detection aims to design a scoring function $S(\rvx)$ that determines whether a given image $\rvx$ has been acquired from the real world (\ie, $S(\rvx)\ge \tau$) or generated by a generative model (\ie, $S(\rvx)<\tau$). While most existing approaches train $S(\rvx)$ using labeled datasets containing both real and synthetic images, we assume a more practical setting in which such training data is not available. This setup allows us to assess the generalizability of detection methods to previously unseen data distributions.

Recently, this generalizability to unseen distributions has been discussed in training-based approaches as well. ZED \cite{Coozolino24MDL} proposes an entropy-based score that can be trained solely on real image distributions without requiring any synthetic examples.  \citet{Coozolino24CLIPfewshot} train a Linear SVM classifier on CLIP \cite{radford2021CLIP} embeddings extracted from image pairs (\eg, MS-COCO \cite{lin2014MSCOCO} and LDM-generated \cite{Rombach22SD}) and evaluate its performance on unseen images (\eg, Raise-1K \cite{Raise2015Nguyen} and Firefly \cite{Adobe25Firefly}).  \citet{rajan2025staypositivecaseignoringreal} improve robustness to post-processed images by retraining the final layer of the pre-trained detection model. Extending beyond these approaches, we focus on a training-free detection setting in which no real or fake data is provided during the design of the detection score.

Training-free detection aims to construct a universal score based on the outputs of a pre-trained foundation model. Such a foundation model may either be a generative model itself (\eg, LDM \cite{Rombach22SD}) or a model trained on diverse real-world images (\eg, DINOv2 \cite{oquab24dinov2}, CLIP \cite{radford2021CLIP}). In this work, we adopt the latter approach, leveraging models pre-trained on real data to guide our detection framework.

\subsection{Self-supervised Models}

Self-supervised models aim to learn robust representations from large-scale unannotated images that can be generalized to a wide range of downstream tasks. Self-supervised learning methods apply various augmentations, often referred to as views, to a single image and train the model to maximize the similarity between outputs corresponding to different views. This paradigm has been implemented in various forms, including contrastive learning (\eg, SimCLR \cite{Chen20SimCLR}, Moco \cite{He20Moco}), clustering-based (\eg, SwaV \cite{Caron20Swav}), and knowledge distillation (\eg, DINO \cite{Caron21Dino}). Since providing a comprehensive overview of self-supervised learning (SSL) is beyond the scope of this paper, we focus on the DINOv2 \cite{oquab24dinov2} model and the augmentation strategies.

DINOv2 is a Vision Transformer \cite{dosovitskiy21ViT} model trained on a web-scale LVD-142M dataset, building upon the iBOT \cite{zhou22IBOT} framework. DINOv2 tokenizes each input image and outputs patch-level embeddings along a \texttt{[CLS]} token embedding that summarizes the entire image. The model is trained using a teacher-student framework, where the student network is optimized to maximize the similarity between its output and the output of the teacher network given relatively mild augmentations. The teacher network is updated via an exponential moving average of the student parameters.

A core component of the data augmentation applied to both networks is \texttt{RandomResizedCrop} (RRC), which randomly crops the region from the input image and resizes it to the given resolution. This operation enforces spatial invariance by encouraging the model to recognize that different subregions of an image correspond to the same underlying semantic content. Due to its effectiveness, RRC has become a standard augmentation technique across many SSL frameworks. In this work, we adopt RRC as the key motivation for designing our AI-generated image detection method.

%% file: sec/3_method.tex
\section{Method} \label{sec:method}

This section introduces our method, \methodname, inspired by the \texttt{RandomResizedCrop} (RRC) operation. We first propose a score function that measures the sensitivity of self-supervised model features to perturbations along high-frequency directions induced by wavelet decomposition (Section~\ref{ssec:wavelet_sensitivity}). Subsequently, we simulate local robustness by introducing a rescaling-and-patching paradigm that partitions resized images into subregions for evaluation (Section~\ref{ssec:rescalepatch_method}).

\input{fig/fig_basescore}

\subsection{Base Detection Score}\label{ssec:wavelet_sensitivity}

We hypothesize that measuring an image's robustness to RRC can serve as an effective score for detecting AI-generated images. However, RRC involves random cropping over a broad hyperparameter space, which introduces variance and may discard key image content of the original image. To address this, we do not apply RRC directly but instead propose a score function that approximates its effect. Specifically, we examine how RRC alters the high-frequency components of the image using wavelet decomposition, motivated by the previous works that AI-generated images exhibit different high-frequency information \cite{schwarz2021frequency,corvi23diffusion}.



For the analysis, we visualize the high-frequency components obtained via Haar wavelet decomposition under multiple instances of RRC, as illustrated in Figure~\ref{subfig:base_rectangle}. Figure~\ref{subfig:base_score_org} presents the high-frequency component within the red-marked region of the original image, whereas Figures~\ref{subfig:base_score_crop1}, \ref{subfig:base_score_crop2}, and \ref{subfig:base_score_crop3} show the corresponding components after applying RRC with green, blue, and yellow cropping regions, respectively. These comparisons reveal that RRC introduces substantial variations in the high-frequency components, even when the cropped region closely matches the original image in size. Based on this observation, we define our base score function as the model’s sensitivity to perturbations along the high-frequency directions, formally expressed as follows:
\begin{equation}
    \text{\basename}(\rvx) =  \frac{\mathbf{f_{M}(\rvx)} \cdot \mathbf{f_{M}(\rvx-\alpha \text{HF}(\rvx))}}{\|\mathbf{f_{M}(\rvx)}\| \|\mathbf{f_{M}(\rvx-\alpha \text{HF}(\rvx))}\|},
\end{equation}
where $\mathbf{f_{\text{M}}}$ denotes the feature output of the self-supervised model $\text{M}$ (e.g., the \texttt{[CLS]} token output of DINOv2), $\text{HF}$ represents the high-frequency component derived from wavelet decomposition, and $0<\alpha<1$ is the perturbation weight. The sensitivity score function $\text{\basename}$ quantifies the change of model $\text{M}$ to perturbations along high-frequency directions. Our central hypothesis is that model $\text{M}$, having been trained on real images, is encouraged to be invariant to such high-frequency perturbations. Accordingly, we expect the $\text{\basename}$ score to be higher for real images than the AI-generated images.

\input{fig/fig_rescalepatch}
\subsection{WaRPAD}\label{ssec:rescalepatch_method}

We further propose a test-time augmentation strategy inspired by RRC. Our main idea is to deterministically simulate multiple instances of RRC by explicitly rescaling the image and thereby patchifying the image to patches with the same sizes as follows:
\begin{equation}
    \texttt{RescaleNPatchify}(\rvx) = \texttt{Patchify}\left(\texttt{Rescale}(\rvx,d_{\text{rescale}}),d_{\text{patch}}\right),    
\end{equation}
where $d_{\text{rescale}}$ and $d_{\text{patch}}$ are the dimension of the rescaled image and the dimension of the patch, $\texttt{Rescale}(\rvx,d)$ is the image rescaling operation to dimension $d \times d$, and $\texttt{Patchify}(\rvx,d)$ is the image patchifying operation to patch dimension $d \times d$,  respectively. 

The \texttt{RescaleNPatchify} operation results in $n_{\text{patch}}=(\frac{d_{\text{rescale}}}{d_{\text{patch}}})^{2}$ number of patches with $d_{\text{patch}}\times d_{\text{patch}}$ dimension. We now examine the proposed sensitivity score in Section \ref{ssec:wavelet_sensitivity} and average the sensitivity score to output the unified score function as follows:
\begin{equation}
    \text{\methodname}(\rvx) = \frac{1}{n_{\text{patch}}}\sum_{\rvx_{\text{patch}}\in\texttt{RescaleNPatchify}(\rvx)} \text{\basename}(\rvx_{\text{patch}})
\end{equation}
Our proposed \methodname examines the sensitivity score in the resized subregion of the images, which are multiple instances of RRC with the area of $\frac{1}{n_{\text{patch}}}$ resized to a dimension of $d_{\text{patch}} \times d_{\text{patch}}$. Since the model $\text{M}$ is trained to be invariant under various RRC instances that include crops similar to the patch, we expect the base score \basename evaluated on the patch to be still robust. On the other hand, We hypothesize the model output of the AI-generated image will deviate in the patches, hence our \methodname further improves on detecting AI-generated images. We verify our hypothesis by examining the \methodname and our base score in 1000 real RAISE-1k \cite{Raise2015Nguyen} data and 1000 SDv1.4-generated data in the Synthbuster benchmark \cite{QuentinSynthbuster}. We select $d_{\text{rescale}}=1344$ and $d_{\text{patch}}=224$.

We show the histogram of real and AI-generated images under our base score on the image, base score on the patch, and the averaged \methodname score in Figure \ref{subfigure:method_rescalepatch_base}, \ref{subfigure:method_rescalepatch_patch}, and \ref{subfigure:method_rescalepatch_mean}, respectively. AI-generated images lose their robustness in our wavelet-based high-frequency perturbation when examined in patches generally. The discrimination between AI-generated images and real images is strengthened in our final score induced by averaging the score function across patches.

%% file: fig/fig_basescore.tex
\begin{figure}[t]
    \centering
    \begin{subfigure}[b]{0.246\textwidth}
        \includegraphics[width=\textwidth]{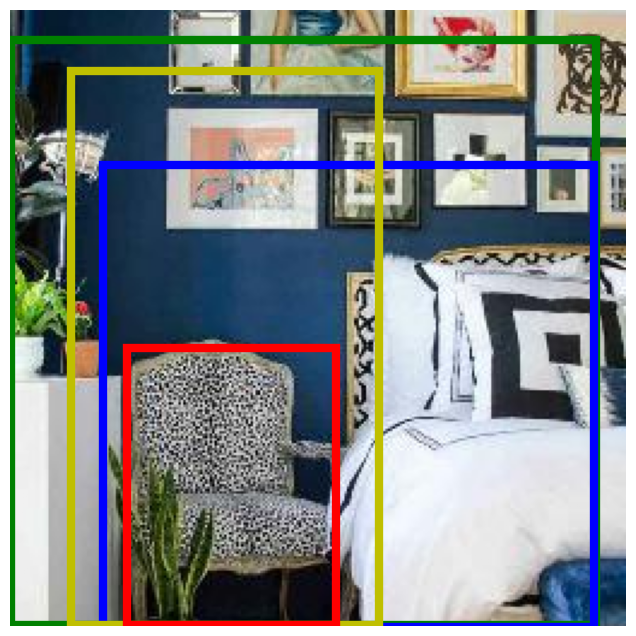}
        \caption{}
        \label{subfig:base_rectangle}
    \end{subfigure}
    \hfill
    \begin{subfigure}[b]{0.182\textwidth}
        \includegraphics[width=\textwidth]{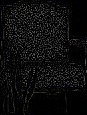}
        \caption{}
        \label{subfig:base_score_org}
    \end{subfigure}
    \hfill    
    \begin{subfigure}[b]{0.182\textwidth}
        \includegraphics[width=\textwidth]{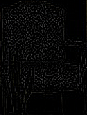}
        \caption{}
        \label{subfig:base_score_crop1}
    \end{subfigure}
    \hfill    
    \begin{subfigure}[b]{0.182\textwidth}
        \includegraphics[width=\textwidth]{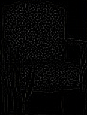}
        \caption{}
        \label{subfig:base_score_crop2}
    \end{subfigure}
    \hfill
    \begin{subfigure}[b]{0.182\textwidth}
        \includegraphics[width=\textwidth]{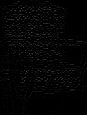}
        \caption{}
        \label{subfig:base_score_crop3}
    \end{subfigure}
    \caption{\textbf{Motivation for the Haar-wavelet perturbation sensitivity score.} \textbf{(a)}: The original image along with the designated region for high-frequency visualization, marked in \textbf{red}. To simulate the effect of \texttt{RandomResizedCrop} (RRC), we apply cropping regions indicated in green, blue, and yellow. \textbf{(b)}: The high-frequency component of the original (uncropped) image obtained via Haar wavelet decomposition. \textbf{(c)}, \textbf{(d)}, \textbf{(e)}: The corresponding high-frequency components of the RRC-transformed images, where the cropping regions are defined by the green, blue, and yellow boxes, respectively. }
    \label{fig:method_wavelet_score}
\end{figure}

%% file: fig/fig_rescalepatch.tex
\begin{figure}[t]
    \centering
    \begin{subfigure}[b]{0.32\textwidth}
        \includegraphics[width=\textwidth]{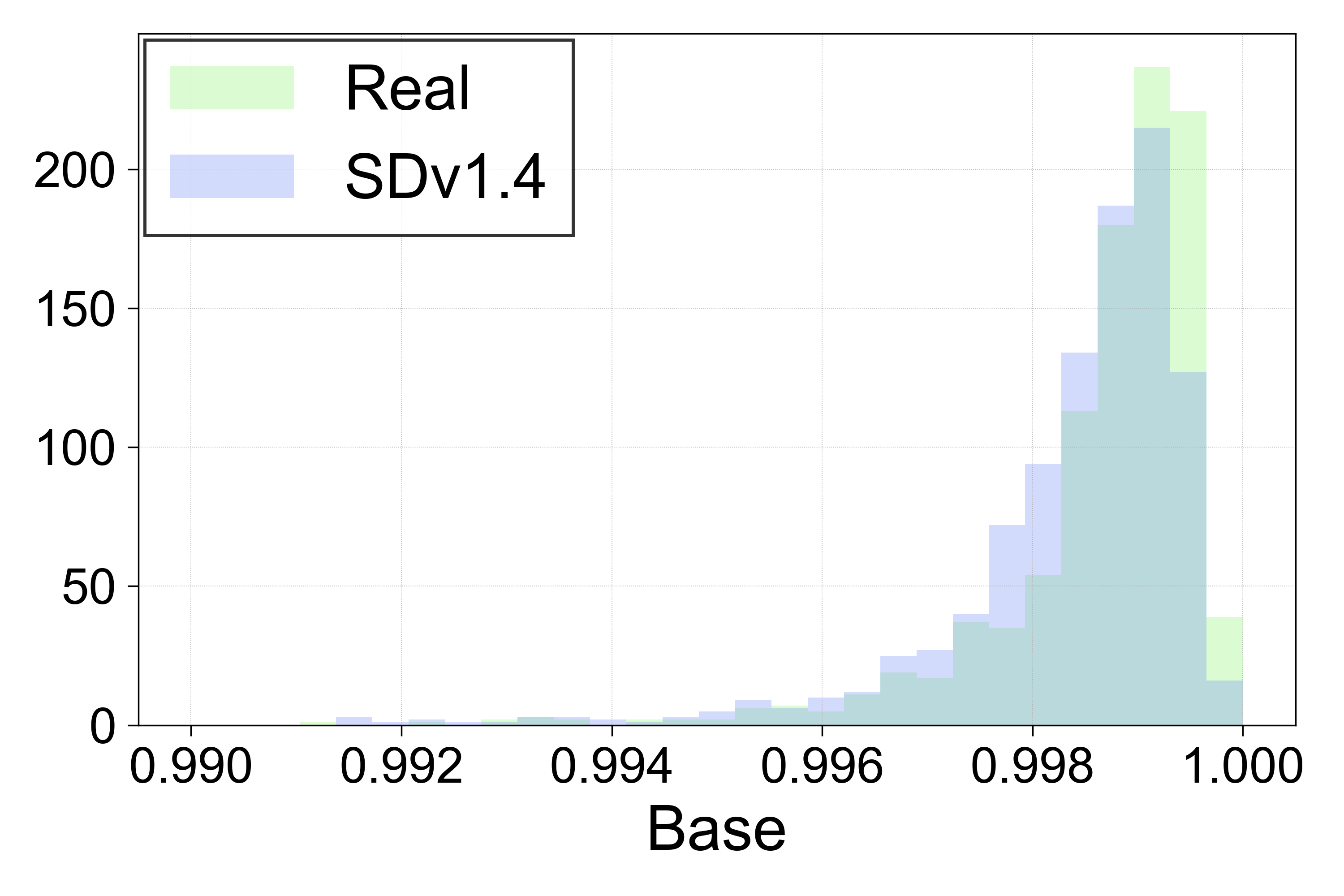}
        \caption{}
        \label{subfigure:method_rescalepatch_base}
    \end{subfigure}
    \hfill
    \begin{subfigure}[b]{0.32\textwidth}
        \includegraphics[width=\textwidth]{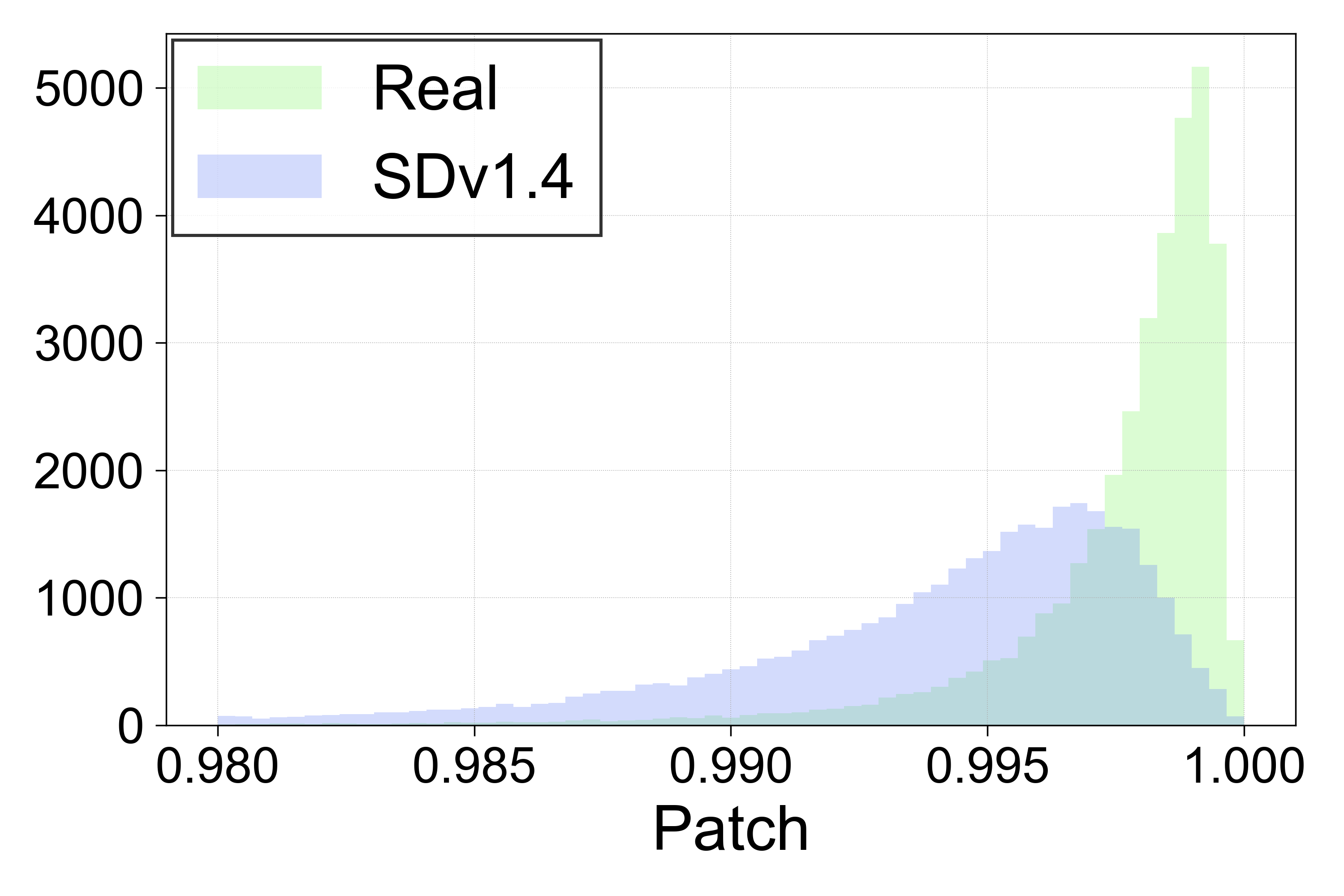}
        \caption{}
        \label{subfigure:method_rescalepatch_patch}
    \end{subfigure}
    \hfill    
    \begin{subfigure}[b]{0.32\textwidth}
        \includegraphics[width=\textwidth]{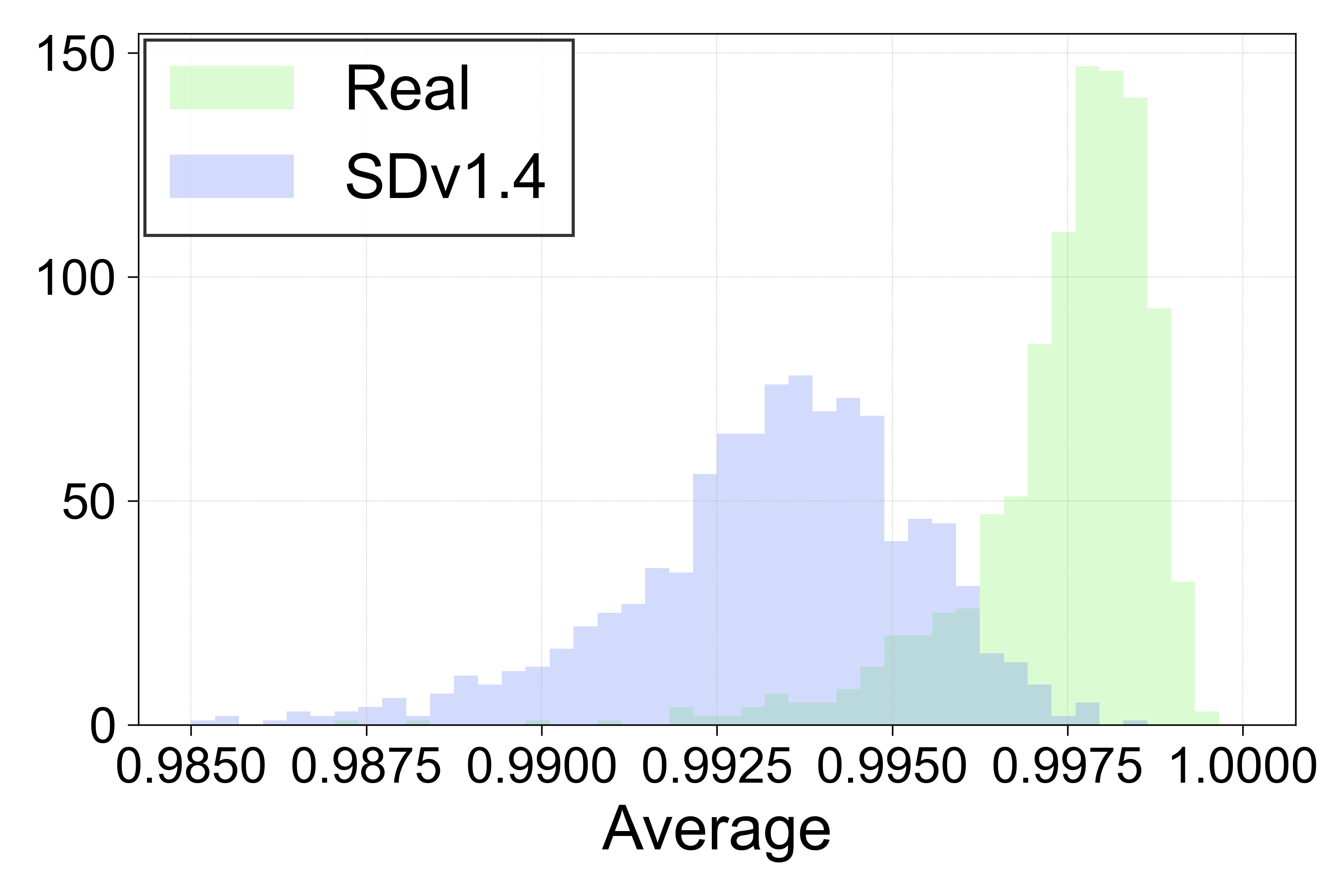}
        \caption{}
        \label{subfigure:method_rescalepatch_mean}
    \end{subfigure}
    \caption{\textbf{Effect of \texttt{RescaleNPatchify}.} \textbf{(a)}: Histogram of real and SDv1.4-generated data examined by \basename. \textbf{(b)}: Histogram of real and SDv1.4-generated data examined on patches augmented by $\texttt{RescaleNPatchify}$. \textbf{(c)}: Histogram of real and SDv1.4-generated data examined on our \methodname score function.}
    \label{fig:method_rescalepatch}
\vspace{-0.1in}
\end{figure}

%% file: sec/4_experiment.tex
\input{tab/table_main_datasets}
\input{tab/table_main_synthbuster}
\input{tab/table_main_genimage}
\input{tab/table_main_lsunbroom}
\input{fig/fig_patchviz}

\section{Experiment}\label{sec:experiment}

We now evaluate \methodname's efficacy on AI-generated image detection. We first introduce the benchmarks and generative models for each benchmark as well as the baseline methods (Section~\ref{ssec:exp_settings}). We report the performance of \methodname across these benchmarks (Section~\ref{ssec:main_results}). Finally, we present detailed ablation studies of \methodname and test its robustness in corruptions (Section~\ref{ssec:ablation_studies}). 

\subsection{Experiment Settings}\label{ssec:exp_settings}

\textbf{Datasets.} We first test \methodname in Synthbuster \cite{QuentinSynthbuster} benchmark based on RAISE-1k dataset \cite{Raise2015Nguyen} where 9 generative models are applied: Firefly \cite{Adobe25Firefly}, GLIDE \cite{Nichol22GLIDE}, SDXL \cite{podell24SDXL}, SDv2, SDv1.3, SDv1.4 \cite{Rombach22SD}, DALL-E 3 \cite{ramesh2023dalle3}, DALL-E 2 \cite{ramesh2022dalle2}, and Midjourney \cite{22Midjourney}. We also test \methodname in GenImage \cite{Zhu23genimage} benchmark where the real data is from the ImageNet \cite{Deng09ImageNet} dataset and 8 generative models are applied for fake image generation: ADM \cite{Dhariwal21ADM}, BigGAN \cite{Brock18Biggan}, GLIDE, Midjourney, SDv1.4, SDv1.5, VQDM \cite{Gu22VQDM}, and Wukong \cite{Gu22Wukong}. Finally, we test on the deepfake-LSUN-bedroom benchmark \cite{Ricker24LSUN} containing 10 generative models: ADM, DDPM \cite{Ho20DDPM}, Diff-ProjectedGAN, Diff-StyleGAN2 \cite{Wang23DiffusionGAN}, IDDPM \cite{Nichol21IDDPM}, LDM \cite{Rombach22SD}, PNDM \cite{Liu22PNDM}, ProGAN \cite{Karras18Progan}, ProjectedGAN \cite{Sauer21ProjectedGAN}, and StyleGAN \cite{Karras19StyleGAN}. We summarize the main information of these benchmarks in Table \ref{table:main_dataset}.

\textbf{Baselines.} We consistently compare against available training-free baselines. AEROBLADE \cite{ricker24aeroblade} and Manifold Induced Bias \cite{Brokman25Manifold} apply SD for the examination. Consistent with the proposal of \citet{Brokman25Manifold}, we apply SDv1.4 for the inspection model. We also compare against RIGID \cite{He24rigid} and MINDER \cite{Tsai24MINDER}, which apply DINOv2 for AI-generated image detection. We follow the original hyperparameter settings from the paper. To further examine the efficacy of training-free setting, we also compare with the leading training-based methods \cite{Yan24Combine, liu2023fatformer} for comparison. Specifically, we examine the pre-trained checkpoint of each method in each benchmark without further training. Note that AIDE is trained on real ImageNet data and SDv1.4-generated data, while FatFormer is trained on LSUN data and ProGAN-generated data.

\textbf{Implementation Details.} The performance of all methods is reported by the area under the ROC curve (AUROC). Consistent with RIGID and MINDER, we use the DINO-ViT-L14 model as the base model. We use the Haar wavelet with a 2-level decomposition to extract the high-frequency information. We also set $d_{\text{patch}}$ and $\alpha$ to 224 and 0.1 throughout all experiments.  For the rescaling dimension $d_{\text{rescale}}$, we set 896 for the GenImage and Deepfake-LSUN-bedroom benchmark and 1344 for the Synthbuster benchmark. We implement our code in the Pytorch \cite{Paskze19Pytorch} framework. All experiments are done on a single A100 GPU.

\subsection{Main Results} \label{ssec:main_results}

Tables~\ref{table:main_synthbuster}, \ref{table:main_genimage}, and \ref{table:main_lsunbroom} report the performance of \methodname compared to other training-free approaches in Synthbuster, GenImage, and Deepfake-LSUN-bedroom benchmark, respectively. \methodname achieves the best performance on all benchmarks on average. Notably, \methodname consistently outperforms RIGID and MINDER by a wide margin of over $6\%$ in AUROC. 

On the other hand, diffusion-model-based methods fail to show consistent performance compared to \methodname. For example, while AEROBLADE is competitive to \methodname in the GenImage benchmark, where some generative models share the same autoencoder with the inspected SDv1.4 model (\eg, SDv1.5, Wukong), its performance deteriorates on detecting proprietary models (\eg, Firefly, DALL-E 2) or GAN-based models. On the other hand, \citet{Brokman25Manifold} can efficiently perform in the Deepfake-LSUN-Bedroom benchmark but fails on the GenImage and Synthbusters benchmarks.  

Finally, training-based methods fail to generalize to unobserved real data distribution and underperform over our \methodname. Specifically, while AIDE performs well on detecting SD-generated data in the GenImage benchmark, it underperforms on other generative models and Deepfake-LSUN-Bedroom/ Synthbuster benchmarks. A similar phenomenon occurs in FatFormer, which shows underwhelming performance in GenImage/Synthbuster benchmarks. On the other hand, \methodname shows the best performance in general, even improving over the FatFormer in LSUN-based benchmark.


We also visualize the patch-wise score information of the real and AI-generated images. Figure~\ref{fig:patch_visualization} shows the patch with the highest \basename score (denoted as \textcolor{red}{red}) and the patch with the lowest score (denoted as \textcolor{blue}{blue}) in the real and AI-generated data in GenImage and Deepfake-LSUN-Bedroom benchmark, respectively. While \basename assigns high scores on patches with rich texture information, the region does not always align with the semantics of the image (\eg, Figure~\ref{subfig:main_patch_real_LSUN}).

\input{tab/table_analysis_ablation}
\input{fig/fig_hyperparameter}
\input{tab/table_analysis_aggregation}
\input{tab/table_analysis_wavelet}
\input{fig/fig_robustness}
\input{tab/table_analysis_backbone}
\input{tab/table_analysis_art}

\subsection{Analysis} \label{ssec:ablation_studies}

We formulate our analysis to validate the following questions:

\begin{itemize}[leftmargin=15pt,topsep=0pt]
\setlength\itemsep{0em}
\item Does each component of \methodname contribute to consistent performance gains? 
\item Is \methodname robust to design choices, hyperparameter ablation, and test-time perturbations?
\item Is \methodname effective for other domains? 
\end{itemize}



\textbf{Effect of each Component.} We first analyze the effect of our proposed \basename and \texttt{RescaleNPatchify} operation. We also investigate the synergy of \basename\xspace and \texttt{RescaleNPatchify} by experimenting with RIGID \cite{He24rigid} with the same $\texttt{RescaleNPatchify}$ procedure. Further, we experiment with the $n_{\text{patch}}$-ensemble version of RIGID that takes the same computational cost as our \methodname. We show the results in Table~\ref{table:ablate} with a comparison against RIGID \cite{He24rigid}. Our perturbation-based metric improves over RIGID in two out of 3 benchmarks. Moreover, RIGID combined with our proposed $\texttt{RescaleNPatchify}$ shows relatively little improvement from RIGID compared to \methodname, highlighting the efficacy of both components.

\textbf{Hyperparameter Analysis.} We analyze the effect of the hyperparameters of the \methodname independently. Figure~\ref{subfigure:hyperparameter_weight} analyzes the effect of perturbation weight $\alpha$ in the Synthbuster benchmark. We also analyze the effect of the rescaling dimension $d_{\text{rescale}}$ on separate benchmarks in Figure~\ref{subfigure:hyperparamer_upscale}. All hyperparameter choices consistently improve over the base dimension, 224. Finally, as DINOv2 can accept arbitrary patch sizes, we also test \methodname in the different patch dimensions $d_{\text{patch}}$. We show the result in Figure~\ref{subfigure:hyperparameter_patch} where the base choice of 224 performs the best. 

\textbf{Design Choices.} We first show the AUROC result of \methodname across different aggregation rules of the computed patch-wise metric and the backbone DINOv2 version. For the aggregation rule, we test the mean, median, minimum, and maximum across patches. For the backbone model, we experiment with 'ViT-S14', 'ViT-B14', 'ViT-L14', and 'ViT-g14'. We also experiment with the 'ViT-L14' and 'ViT-g14' with register tokens \cite{Darcet24Register}. 

We show the result in Table~\ref{table:aggregation}. Note that Mean and 'ViT-L14' correspond to the results in Section~\ref{ssec:main_results}. In the perspective of the aggregation rule, the mean or median aggregation rule achieves the best performance consistently. On the other hand, concerning the DINOv2 backbone, a larger model size shows better results with the slight exception of the 'ViT-L14' backbone in the GenImage benchmark. 

We further experiment with the different choices of the wavelet and the decomposition level of the wavelet decomposition. Apart from the Haar wavelet, we experiment with Daubeches wavelets (db2, db3, db4), Biorthogonal wavelets (bior1.3, bior1.5, bior2.2, bior2.4, bior3.1), and Coiflet wavelets (coif1, coif2, coif3) with decomposition levels from 1 to 3. 

We present the results in Table~\ref{table:wavelet}, grouping wavelets by the number of vanishing moments in the (synthesis) wavelet function $\psi$. While the Haar wavelet achieves the highest performance, our results indicate that other wavelets with one vanishing moment also perform competitively. In contrast, wavelets with higher vanishing moments tend to induce more structured perturbations on the real images, and we find that DINOv2 is no longer robust to the perturbation. We further report this phenomenon in the Appendix, where we include histograms of the score distributions for other wavelets for both real and AI-generated images.


\textbf{Robustness to Corruptions.} Our \methodname is based on the self-supervised vision model, which may learn robust representations due to training on a wide range of perturbations (\eg, \texttt{ColorJitter}, \texttt{RandomSolarize}). Hence, we test \methodname's robustness on the corruption of the input images, both real and AI-generated. For comparison, we also test RIGID, MINDER, and AEROBLADE on the same corruption. We report the average performance on the GenImage benchmark with JPEG compression, 
center crop and resizing, and Gaussian noise corruptions with varying degrees. 

Figure~\ref{fig:robustness_genimage} shows the performance of \methodname as well as the training-free baselines under corruption in the GenImage benchmark. \methodname achieves competitive performance over the baseline in every corruption consistently. On the other hand, AEROBLADE's performance quickly degrades compared to \methodname in high-level corruption. Experiments in other benchmarks in the Appendix exhibit consistent behaviors.

\textbf{Extension to other backbones.} The result in DINOv2 shows that our proposed metric can perform not only in in-domain datasets (\eg, ImageNet) but also in datasets unobserved in the training phase (\eg, RAISE-1k, LSUN-Bedroom). We further experiment \methodname with other backbones of self-supervised models trained to be invariant under RRC. For the model, we select CLIP \cite{radford2021CLIP}, SwaV \cite{Caron20Swav}, and DINO \cite{Caron21Dino}. We also test other models that use RRC only as data augmentation: BeiT \cite{bao2022beit} and ViTMAE \cite{he2022vitmae}. We also specify the details of the tested models in the Appendix. We further include MINDER and RIGID for each backbone as a comparison. We do not change any hyperparameters. 

As shown in Table~\ref{table:backbone}, our \methodname consistently outperforms RIGID and MINDER when the backbone model is trained to be invariant under RRC. However, the gain of \methodname is less prominent when tested in the masked-image-modeling-based backbones and even underperforms over RIGID when the backbone model is BeiT.


\textbf{Examination on other domain data.} While we thoroughly evaluate \methodname on various benchmarks, such benchmarks are from the natural image domain. Motivated by the recent practice \cite{Jia25Color} that evaluates AI-generated image detectors outside the natural image domain, we further test the performance of \methodname on the Art domain. Since the FakeART benchmark in \cite{Jia25Color} is not public except for the information that they use the WikiArt\footnote{\url{https://wikiart.org}} dataset for the real data, we instead download data from the Kaggle webpage\footnote{\url{https://www.kaggle.com/datasets/doctorstrange420/real-and-fake-ai-generated-art-images-dataset}} with 10821/10821 real and AI-generated data available, where the real data is from the WikiArt dataset. For comparison, we evaluate our method against FatFormer, RIGID, and MINDER on the Art domain, and report the results in Table~\ref{table:art}. This demonstrates the effectiveness of \methodname under distribution shift.




%% file: tab/table_main_datasets.tex
\begin{table*}[t!]
\centering
\caption{\textbf{Benchmarks in the main experiment.}}
\label{table:main_dataset}
\resizebox{\textwidth}{!}{
\begin{tabular}{ccccc}
\toprule
                                 
Benchmark & Real Dataset & \# of Generative Models & Input Resolution & \# of Test Images per Dataset\\ 
\midrule
Synthbuster \cite{QuentinSynthbuster} & Raise-1K \cite{Raise2015Nguyen} & 9 (LDM, Proprietary Model) & Varying ($256 \times 256$ $\sim$ $4928 \times 3264$) & 1000 \\
GenImage \cite{Zhu23genimage} & ImageNet \cite{Deng09ImageNet} & 8 (GAN, Diffusion Model, LDM) & Varying ($128 \times 128$ $\sim$ $1024 \times 1024$) & 6000 $\sim$ 8000\\
Deepfake-LSUN-Bedroom \cite{Ricker24LSUN} & LSUN \cite{Yu15LSUN} & 10 (GAN, Diffusion Model) & $256 \times 256$ & 10000 \\
\bottomrule
\end{tabular}}
\vspace{-0.1in}
\end{table*}

%% file: tab/table_main_synthbuster.tex
\begin{table*}[t]
\centering
\caption{\textbf{AI-generated image detection performance (AUROC) of \methodname and baselines in the Synthbuster \citep{QuentinSynthbuster} benchmark.} \textbf{Bold} and \underline{underline} denotes the best method and the second best methods, respectively.}
\label{table:main_synthbuster}
\resizebox{\textwidth}{!}{
\begin{tabular}{ccccccccccc}
\toprule
                                 
Method  & Firefly & GLIDE & SDXL & SDv2 & SDv1.3 & SDv1.4 & DALL-E 3 & DALL-E 2 & Midjourney & Mean \\ 

\midrule
\multicolumn{11}{c}{Training-based Methods} \\ 
\midrule
AIDE \cite{Yan24Combine} & 0.165 & 0.780 & 0.835 & 0.642 & 0.946 & 0.933 & 0.415 & 0.426 & 0.688 & 0.648 \\
FatFormer \cite{liu2023fatformer} & 0.586 & 0.718 & 0.707 & 0.513 & 0.486 & 0.500 & 0.186 & 0.571 & 0.374 & 0.516 \\
\midrule
\multicolumn{11}{c}{Training-free Methods} \\ 
\midrule
$\text{RIGID}$ \citep{He24rigid} & 0.519 & 0.868 & \underline{0.757} & 0.615 & 0.448 & 0.446 & \underline{0.442} & 0.596 & 0.593 & 0.587 \\
$\text{MINDER}$ \citep{Tsai24MINDER} & 0.440 & 0.568 & 0.472 & 0.721 & 0.656 & 0.668 & 0.346 & 0.445 & 0.345 & 0.518 \\
$\text{AEROBLADE}$ \citep{ricker24aeroblade} & \underline{0.592} & \underline{0.954} & 0.668 & 0.567 & \underline{0.950} & \underline{0.950} & \textbf{0.486} & 0.392 & \textbf{0.769} & \underline{0.703} \\
$\text{Manifold Bias}$ \citep{Brokman25Manifold} & 0.493 & 0.779 & 0.562 & \underline{0.749} & 0.544 & 0.549 & 0.379 & \underline{0.607} & 0.424 & 0.565 \\
$\methodname$ \textbf{(ours)} & \textbf{0.927} & \textbf{0.999} & \textbf{0.830} & \textbf{0.775} & \textbf{0.959} & \textbf{0.958} & 0.422 & \textbf{0.930} & \underline{0.702} & \textbf{0.834} \\
\bottomrule
\end{tabular}}
\end{table*}

%% file: tab/table_main_genimage.tex
\begin{table*}[t]
\centering
\caption{\textbf{AI-generated image detection performance (AUROC) of \methodname and baselines in the GenImage \citep{Zhu23genimage} benchmark.} \textbf{Bold} and \underline{underline} denotes the best and second best methods, respectively.}
\label{table:main_genimage}
\resizebox{\textwidth}{!}{
\begin{tabular}{cccccccccc}
\toprule
                                 
Method  & ADM & BigGAN & GLIDE & Midjourney & SDv1.4 & SDv1.5 & VQDM & Wukong & Mean \\ 
\midrule 
\multicolumn{10}{c}{Training-based Methods} \\ 
\midrule
AIDE \cite{Yan24Combine} & 0.921 & 0.920 & 0.987 & 0.959 & 1.000 & 1.000 & 0.965 & 1.000 & 0.969 \\
FatFormer \cite{liu2023fatformer} & 0.903 & 0.995 & 0.951 & 0.579 & 0.780 & 0.776 & 0.967 & 0.824 & 0.847 \\
\midrule
\multicolumn{10}{c}{Training-free Methods} \\ 
\midrule

$\text{RIGID}$ \citep{He24rigid} & \underline{0.874} & 0.974 & 0.952 & 0.778 & 0.682 & 0.682 & \underline{0.915} & 0.699 & 0.820 \\
$\text{MINDER}$ \citep{Tsai24MINDER} & 0.768 & 0.681 & 0.582 & 0.450 & 0.607 & 0.596 & 0.882 & 0.676 & 0.655 \\
$\text{AEROBLADE}$ \citep{ricker24aeroblade} & 0.856 & \underline{0.981} & \underline{0.989} & \textbf{0.918} & \textbf{0.982} & \textbf{0.984} & 0.732 & \textbf{0.983} & \underline{0.928}\\
$\text{Manifold Bias}$ \citep{Brokman25Manifold} & 0.727 & 0.925 & 0.852 & 0.510 & 0.675 & 0.673 & 0.874 & 0.653 & 0.736 \\
$\methodname$ \textbf{(ours)} & \textbf{0.986} & \textbf{0.998} & \textbf{0.991} & \underline{0.810} & \underline{0.940} & \underline{0.936} & \textbf{0.981} & \underline{0.924} & \textbf{0.946} \\
\bottomrule
\end{tabular}}
\end{table*}

%% file: tab/table_main_lsunbroom.tex
\begin{table*}[t!]
\centering
\caption{\textbf{AI-generated image detection performance (AUROC) of \methodname and baselines in the Deepfake-LSUN-Bedroom \citep{Ricker24LSUN} benchmark.} \textbf{Bold} and \underline{underline} denotes the best method and the second best methods, respectively.}
\label{table:main_lsunbroom}
\resizebox{\textwidth}{!}{
\begin{tabular}{cccccccccccc}
\toprule
                                 
Method & ADM & DDPM & Diff-ProjectedGAN & Diff-StyleGAN2 & IDDPM & LDM & PNDM & ProGAN & ProjectedGAN & StyleGAN & Mean \\ 

\midrule
\multicolumn{12}{c}{Training-based Methods} \\ 
\midrule
AIDE \cite{Yan24Combine} & 0.636 & 0.722 & 0.860 & 0.951 & 0.679 & 0.807 & 0.941 & 0.899 & 0.910 & 0.840 & 0.825 \\
FatFormer \cite{liu2023fatformer} & 0.745 & 0.709 & 0.998 & 1.000 & 0.824 & 0.944 & 0.999 & 1.000  & 0.999 & 0.988 & 0.921 \\
\midrule
\multicolumn{12}{c}{Training-free Methods} \\ 
\midrule
$\text{RIGID}$ \citep{He24rigid} & 0.742 & 0.887 & 0.937 & 0.914 & 0.855 & 0.846 & 0.843 & 0.957 & 0.944 & 0.681 & 0.861 \\
$\text{MINDER}$ \citep{Tsai24MINDER} & 0.706 & 0.796 & \underline{0.973} & 0.942 & 0.782 & 0.844 & \underline{0.896} & 0.970 & 0.973 & 0.805 & 0.869 \\
$\text{AEROBLADE}$ \citep{ricker24aeroblade} & 0.545 & 0.741 & 0.488 & 0.534 & 0.656 & 0.595 & 0.382 & 0.454 & 0.490 & 0.342 & 0.522 \\
$\text{Manifold Bias}$ \citep{Brokman25Manifold} & \textbf{0.788} & \underline{0.905} & 0.968 & \underline{0.943} & \underline{0.888} & \underline{0.928} & 0.891 & \textbf{0.996} & \underline{0.978} & \textbf{0.912} & \underline{0.920}\\

$\methodname$ \textbf{(ours)} & \underline{0.785} & \textbf{0.937} & \textbf{0.988} & \textbf{0.965} & \textbf{0.908} & \textbf{0.940} & \textbf{0.970} & \underline{0.995} & \textbf{0.986} & \underline{0.870} & \textbf{0.934} \\
\bottomrule
\end{tabular}}
\vspace{-0.2in}
\end{table*}

%% file: fig/fig_patchviz.tex
\begin{figure}[t]
    \centering
    \begin{subfigure}[b]{0.23\textwidth}
        \includegraphics[width=\textwidth]{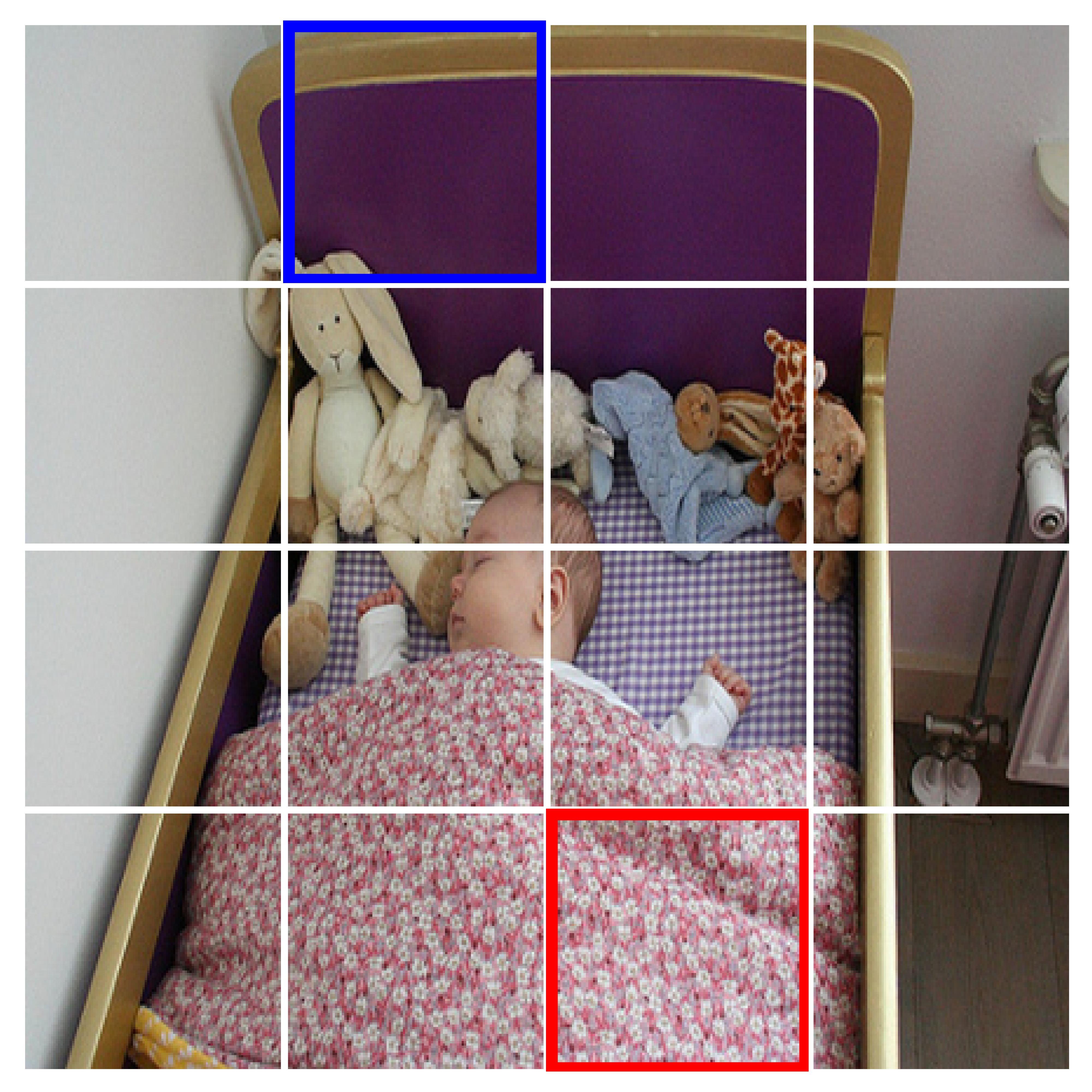}
        \caption{Real-GenImage}
    \end{subfigure}
    \hfill
    \begin{subfigure}[b]{0.23\textwidth}
        \includegraphics[width=\textwidth]{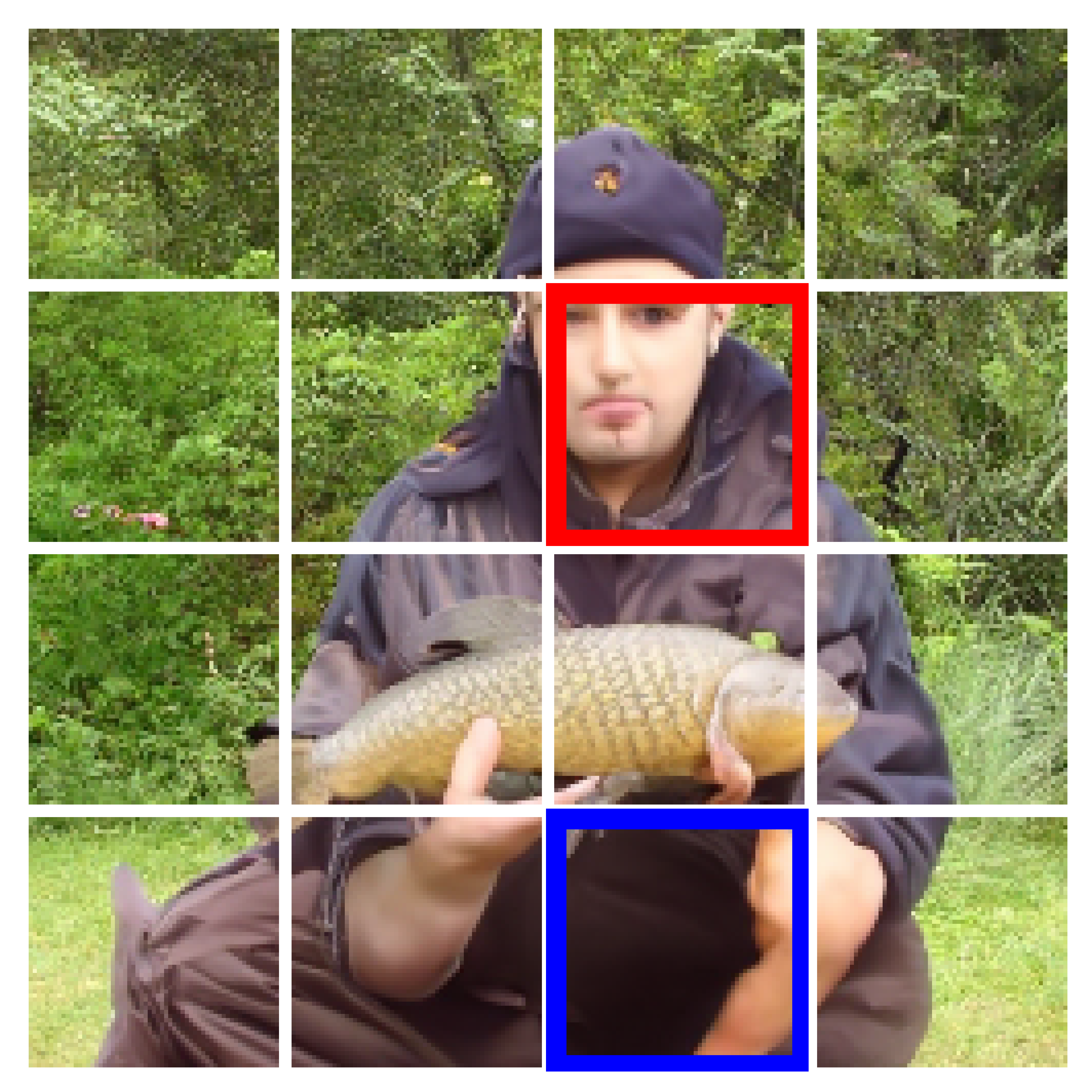}
        \caption{AI-GenImage}
    \end{subfigure}
    \hfill    
    \begin{subfigure}[b]{0.23\textwidth}
        \includegraphics[width=\textwidth]{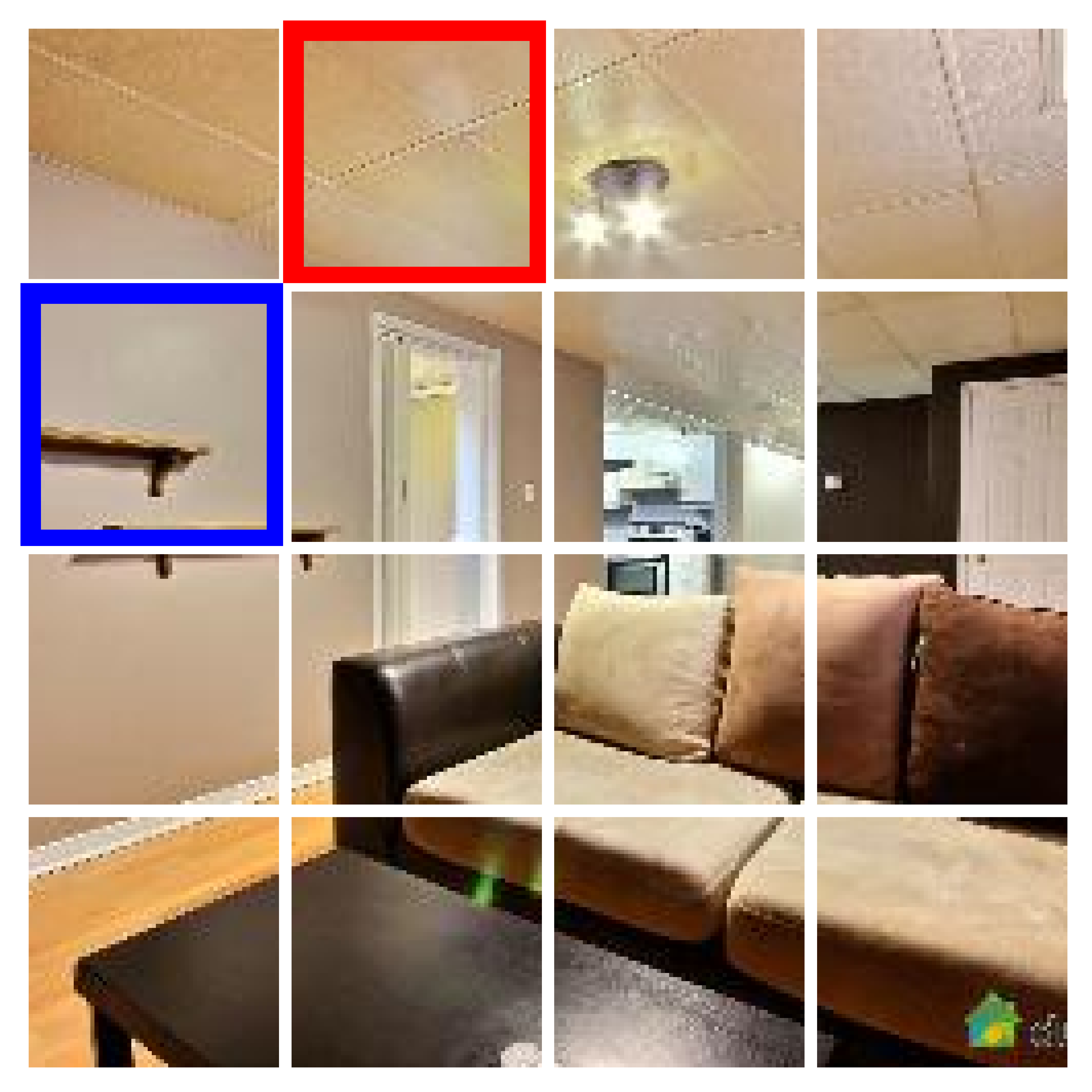}
        \caption{Real-LSUN}
        \label{subfig:main_patch_real_LSUN}
    \end{subfigure}
    \hfill    
    \begin{subfigure}[b]{0.23\textwidth}
        \includegraphics[width=\textwidth]{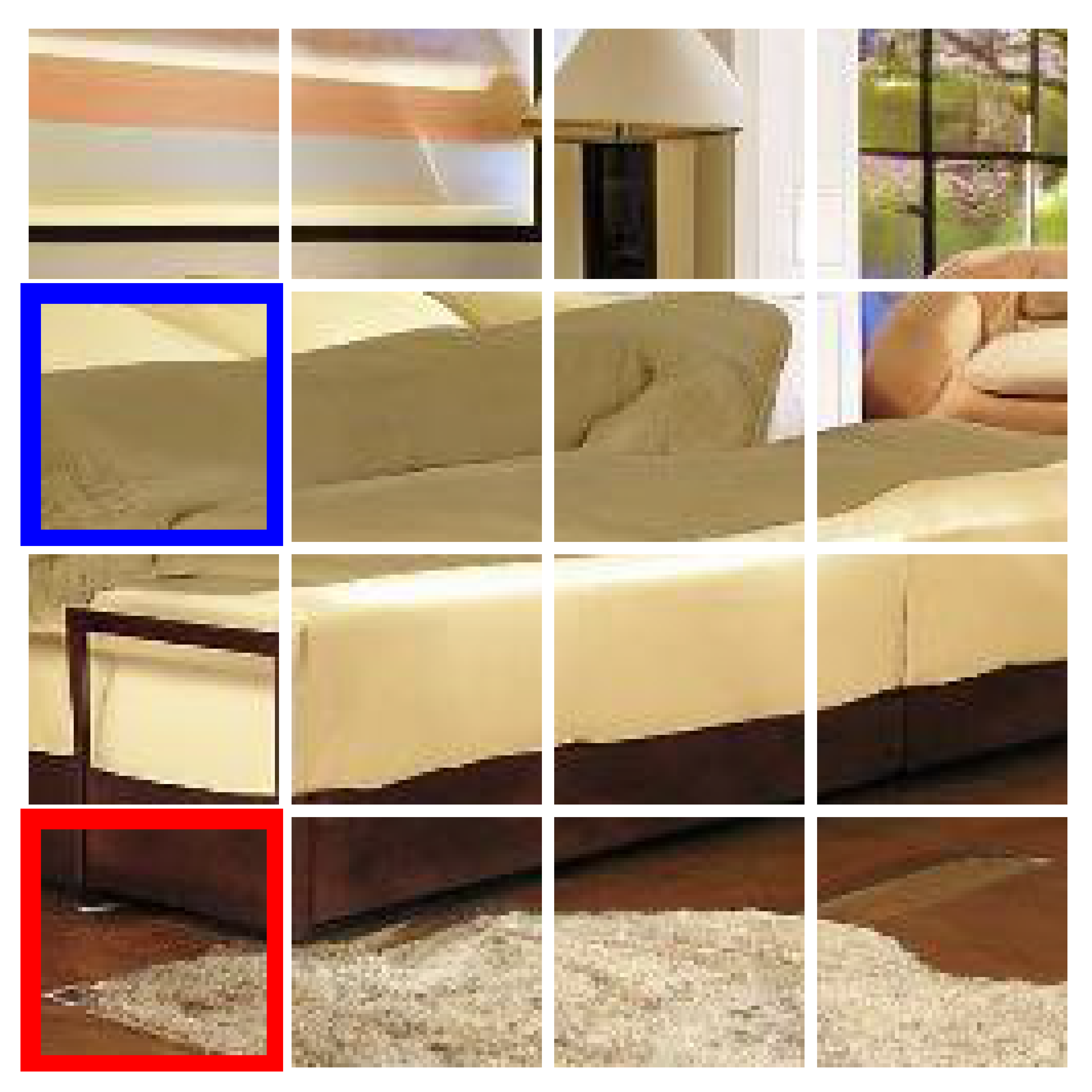}
        \caption{AI-LSUN}
    \end{subfigure}
    \caption{\textbf{Visualization of the \basename score across patches.} We show the patch with the highest score in \textcolor{red}{red} and lowest score in \textcolor{blue}{blue}. Each image is from ImageNet \textbf{(a)}, ADM-generated GenImage \textbf{(b)}, LSUN \textbf{(c)}, and ADM-generated Deepfake-LSUN-Bedroom dataset \textbf{(d)}, respectively.}
    \label{fig:patch_visualization}
\end{figure}

%% file: tab/table_analysis_ablation.tex
\begin{table}[t]
\centering
\caption{\textbf{Ablation Study on each component of \methodname.} We report AUROC. Gains are computed against RIGID \cite{He24rigid}.}
\label{table:ablate}
\resizebox{0.85\textwidth}{!}{%
\begin{tabular}{llll}
\toprule   
Method & Synthbuster & GenImage & Deepfake-LSUN-Bedroom \\
\midrule
RIGID \cite{He24rigid} & 0.587 & 0.820 & 0.861  \\ 
\midrule
RIGID $\times n_{\text{patch}}$ & 0.589 \textcolor{green}{(+0.2\%)} & 0.823 \textcolor{green}{(+0.3\%)} & 0.872 \textcolor{green}{(+1.1\%)} \\
\basename & 0.636 \textcolor{red}{(+4.9\%)} & 0.809 \textcolor{red}{(-1.1\%)} & 0.890 \textcolor{red}{(+2.9\%)}\\
RIGID + \texttt{RescaleNPatchify} & 0.656 \textcolor{blue}{(+6.9\%)} & 0.800 \textcolor{blue}{(-2.0\%)} & 0.861 \textcolor{blue}{(+0.0\%)} \\
\methodname & 0.834 \textcolor{brown}{(+24.7\%)} & 0.946 \textcolor{brown}{(+12.6\%)} & 0.934 \textcolor{brown}{(+7.3\%)} \\
\bottomrule
\end{tabular}}
\end{table} 

%% file: fig/fig_hyperparameter.tex
\begin{figure}[t]
    \centering
    \begin{subfigure}[b]{0.32\textwidth}
        \includegraphics[width=\textwidth]{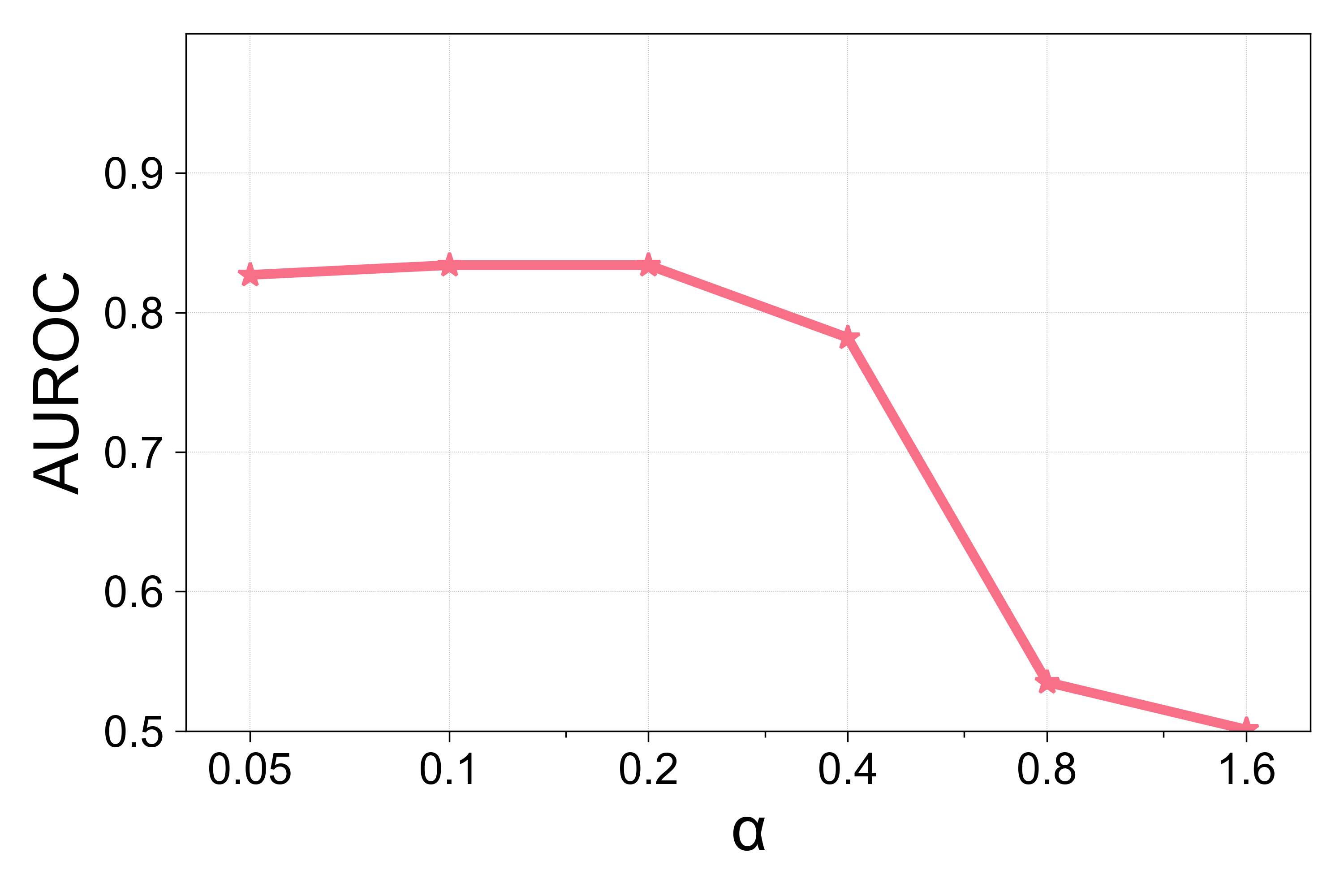}
        \caption{}
        \label{subfigure:hyperparameter_weight}
    \end{subfigure}
    \hfill
    \begin{subfigure}[b]{0.32\textwidth}
        \includegraphics[width=\textwidth]{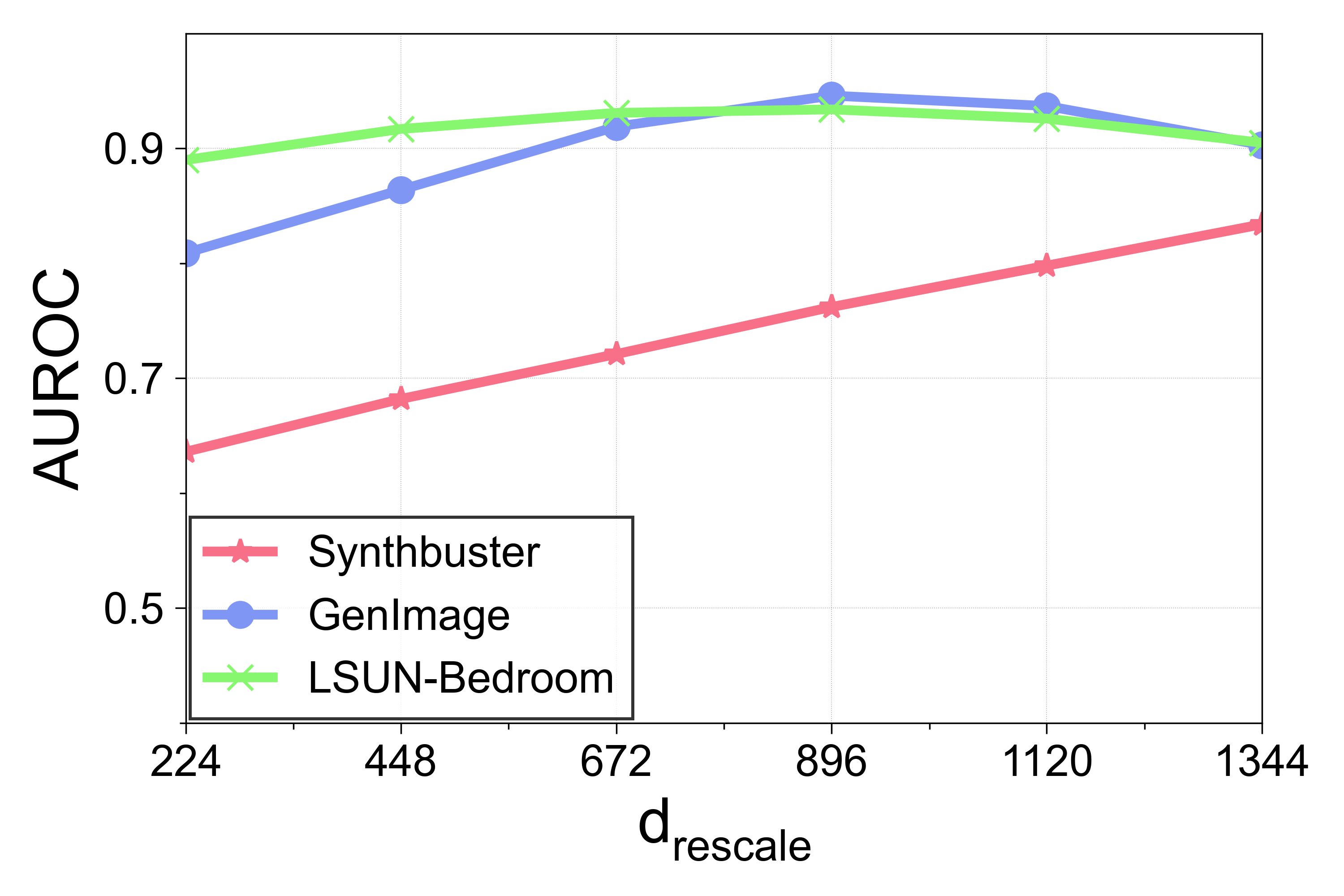}
        \caption{}
        \label{subfigure:hyperparamer_upscale}
    \end{subfigure}
    \hfill    
    \begin{subfigure}[b]{0.32\textwidth}
        \includegraphics[width=\textwidth]{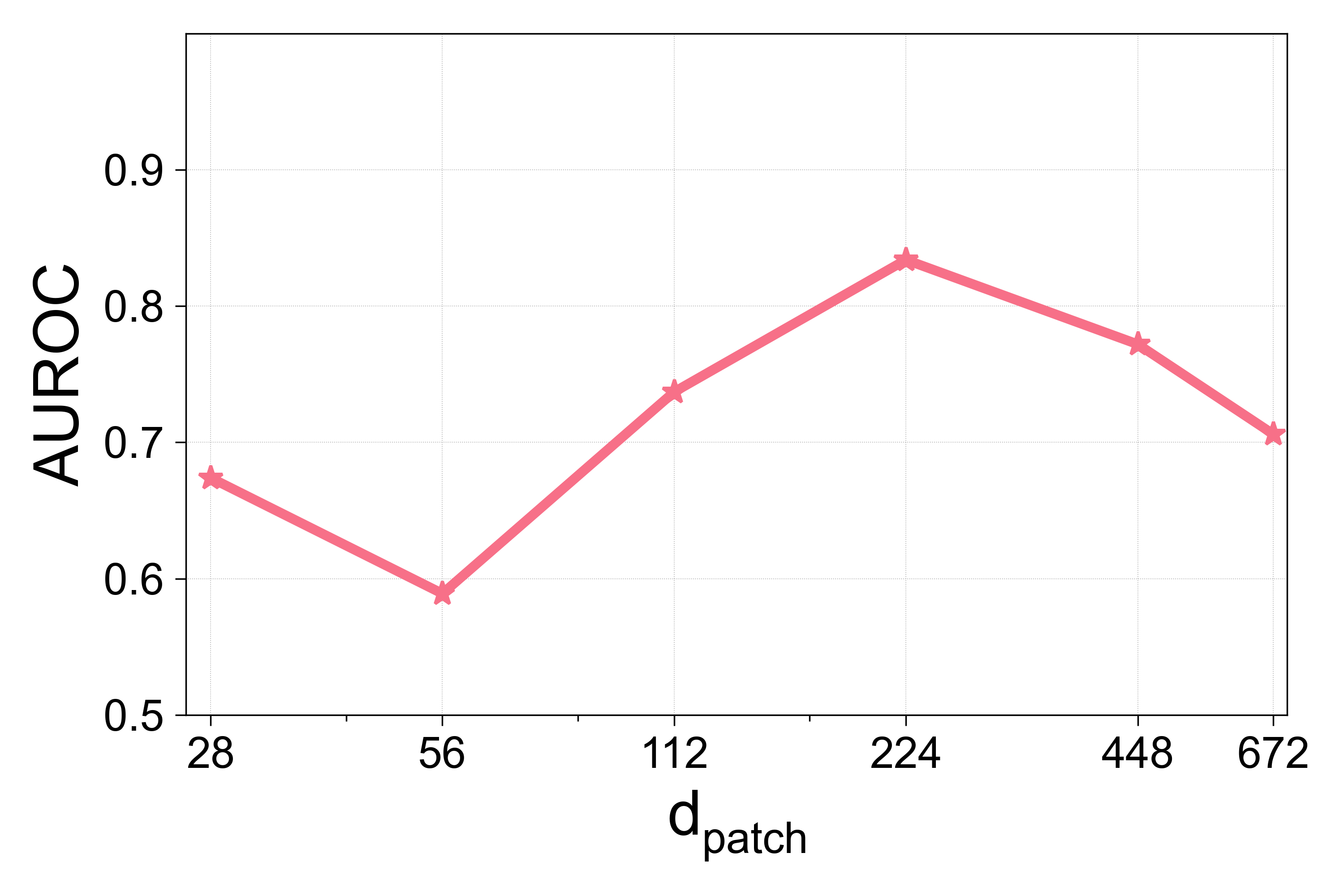}
        \caption{}
        \label{subfigure:hyperparameter_patch}
    \end{subfigure}
    \caption{\textbf{Hyperparameter analysis of \methodname} \textbf{(a)}: AUROC performance of \methodname with respect to $\alpha$ in the Synthbuster benchmark. \textbf{(b)}: AUROC result with respect to the $d_{\text{rescale}}$. \textbf{(c)}: AUROC result with respect to the $d_{\text{patch}}$ in the Synthbuster benchmark.}
    \label{fig:hyperparameter_analysis}
\vspace{-0.15in}
\end{figure}

%% file: tab/table_analysis_aggregation.tex
\begin{table}[t]
\centering
\caption{\textbf{Ablation Study on the DINOv2 backbone and patch-wise aggregation rule on \methodname.} We report AUROC. \textbf{Bold} denotes the best choice.}
\label{table:aggregation}
\resizebox{\textwidth}{!}{%
\begin{tabular}{lllllll}
\toprule   
Agg Rule & ViT-S14 & ViT-B14 & ViT-L14  & ViT-L14-reg & ViT-g14 & ViT-g14-reg  \\
\midrule
Mean & 0.76/0.86/0.83 & 0.82/0.92/0.90 & 0.83/\textbf{0.95}/0.93 & 0.84/0.94/0.94 & 0.86/0.94/\textbf{0.95} & \textbf{0.87}/0.94/\textbf{0.95} \\
Median & 0.77/0.85/0.81 & 0.83/0.91/0.89 & 0.84/0.94/0.92 & 0.85/0.94/0.93 & 0.85/0.93/\textbf{0.95} & \textbf{0.87}/0.94/\textbf{0.95} \\
Min & 0.71/0.85/0.79 & 0.75/0.89/0.84 & 0.76/0.91/0.88 & 0.77/0.91/0.89 & 0.80/0.90/0.91& 0.80/0.90/0.90 \\
Max & 0.68/0.74/0.63 & 0.77/0.81/0.77 & 0.78/0.90/0.79 & 0.78/0.90/0.76 & 0.79/0.90/0.87 & 0.80/0.91/0.87 \\
\bottomrule
\end{tabular}}
\end{table}

%% file: tab/table_analysis_wavelet.tex
\begin{table*}[t!]
\centering
\caption{\textbf{AI-generated image detection performance (AUROC) to wavelet choice and decomposition level in the Synthbuster benchmark.} \textbf{Bold} and \underline{underline} denotes the best and the second best choice. We group the wavelets by the number of vanishing moments on a wavelet function $\psi$.}
\label{table:wavelet}
\resizebox{\textwidth}{!}{%
\begin{tabular}{ccccccccccccc}
\toprule
& \multicolumn{3}{c}{1 vanishing moment} & \multicolumn{9}{c}{>1 vanishing moments} \\
\cmidrule(lr){2-4}\cmidrule(lr){5-13} 
                                 

Level & Haar & bior 1.3 & bior 1.5 & db2 & db3 & db4 & bior 2.2 & bior 2.4 & bior 3.1 & coif1 & coif2 & coif3 \\


\midrule

1 & 0.745 & 0.715 & 0.701 & 0.458 & 0.505 & 0.527 & 0.474 & 0.483 & 0.486 & 0.480 & 0.481 & 0.487 \\

2 & \textbf{0.834} & 0.591 & \underline{0.827} & 0.512 & 0.491 & 0.508 & 0.512 & 0.487 & 0.472 & 0.533 & 0.506 & 0.501 \\

3& 0.787 & 0.585 & 0.569 & 0.466 & 0.490 & 0.460 & 0.476 & 0.490 & 0.486 & 0.498 & 0.457 & 0.467 \\

\bottomrule
\end{tabular}}
\end{table*}

%% file: fig/fig_robustness.tex
\begin{figure}[t]
    \centering
    \begin{subfigure}[b]{0.32\textwidth}
        \includegraphics[width=\textwidth]{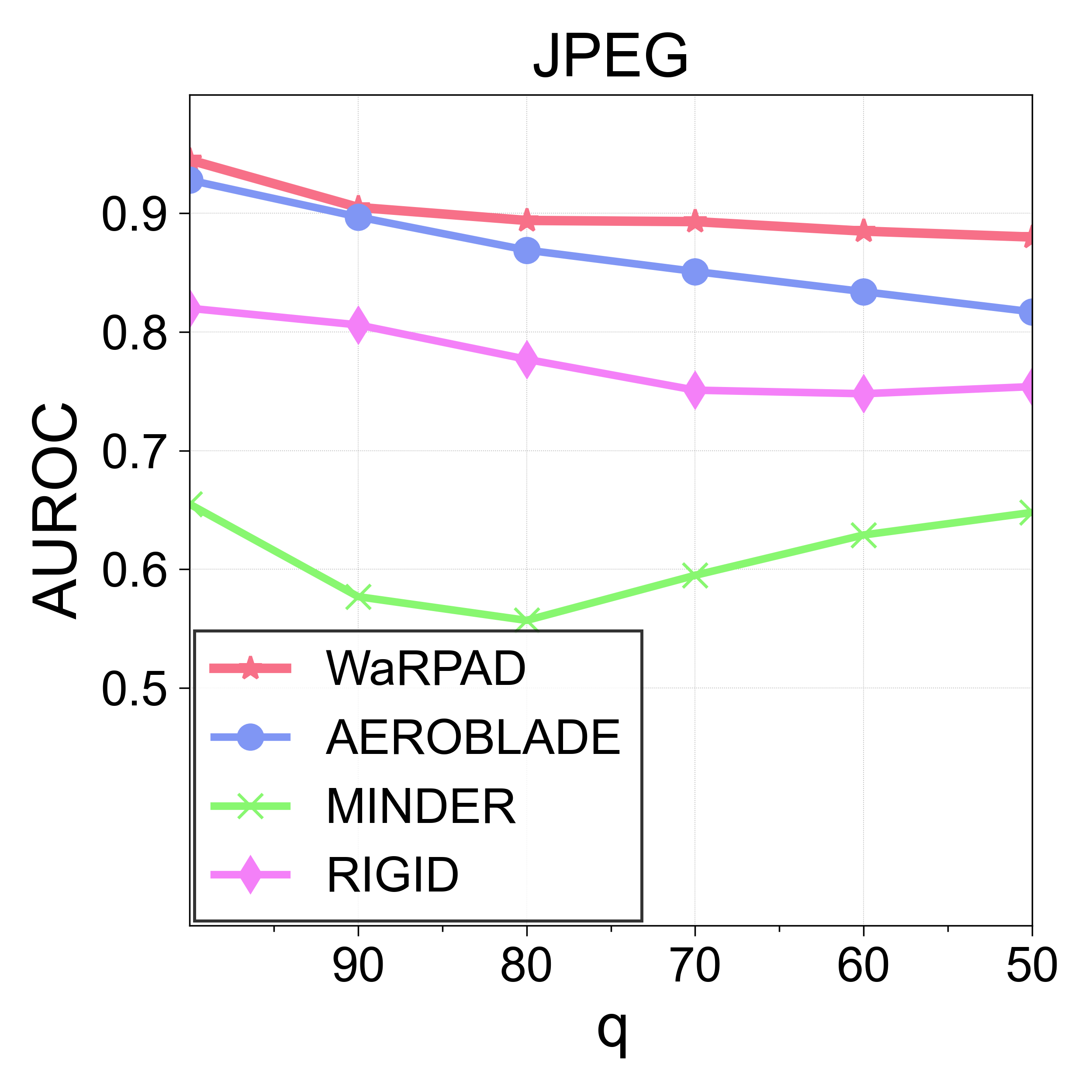}
        \caption{JPEG Compression}
    \end{subfigure}
    \hfill
    \begin{subfigure}[b]{0.32\textwidth}
        \includegraphics[width=\textwidth]{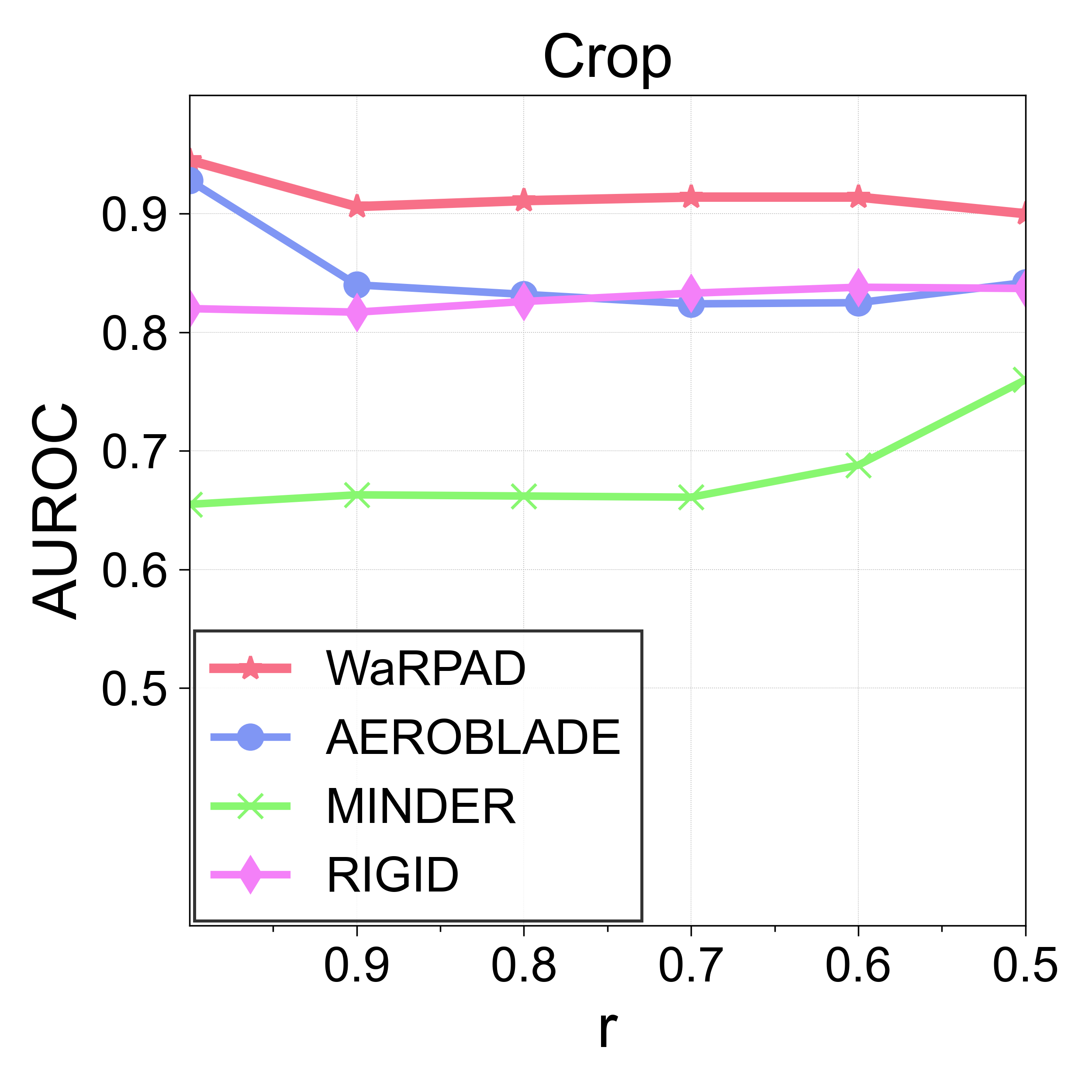}
        \caption{Center Crop}
    \end{subfigure}
    \hfill     
    \begin{subfigure}[b]{0.32\textwidth}
        \includegraphics[width=\textwidth]{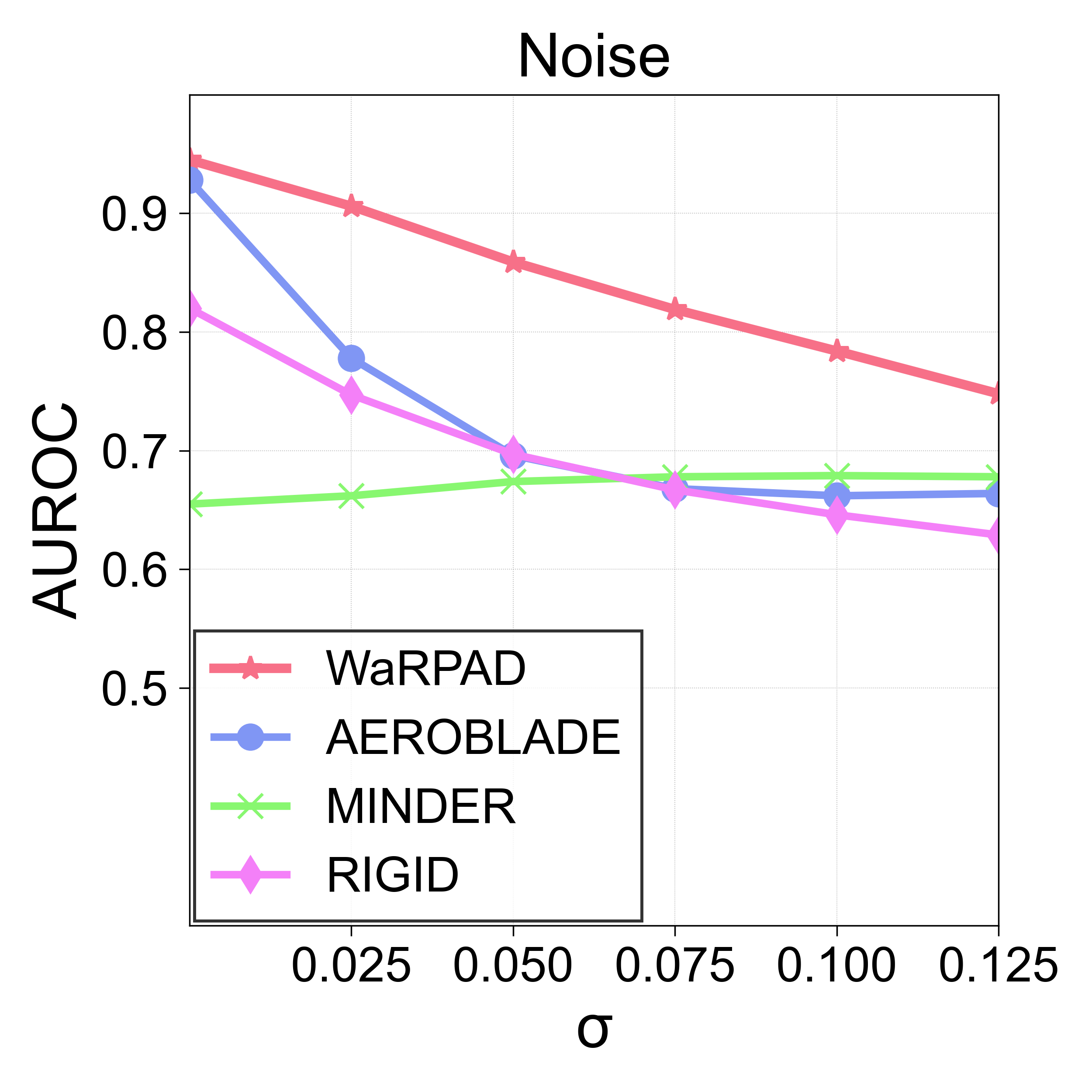}
        \caption{Gaussian Noise}
    \end{subfigure}
    \caption{\textbf{Robustness of \methodname in corruptions.} We test the AUROC performance of \methodname, AEROBLADE, MINDER, and RIGID in corrupted test images in the GenImage benchmark.}
    \label{fig:robustness_genimage}
\vspace{-0.1in}
\end{figure}

%% file: tab/table_analysis_backbone.tex
\begin{table*}[ht!]
\centering
\caption{\textbf{AI-generated image detection performance (AUROC) in the Synthbuster benchmark under different backbones.}}
\label{table:backbone}
\resizebox{\textwidth}{!}{
\begin{tabular}{ccccccc}
\toprule
  & \multicolumn{4}{c}{Invariant to RRC}
  & \multicolumn{2}{c}{Others} \\
  \cmidrule(lr){2-5}\cmidrule(lr){6-7} 
Method & DINOv2 \cite{oquab24dinov2} & CLIP \cite{radford2021CLIP} & SwaV \cite{Caron20Swav} & DINO \cite{Caron21Dino} & BeiT \cite{bao2022beit}  & ViTMAE \cite{he2022vitmae} \\

\midrule
RIGID & 0.587 & 0.561 & 0.542 & 0.480 & 0.619 & 0.531\\
MINDER & 0.518 & 0.583 & 0.542 & 0.478 & 0.473 & 0.545\\
\methodname & 0.834 & 0.802 & 0.743 & 0.707 & 0.486 & 0.620\\
\bottomrule
\end{tabular}}
\vspace{-0.1in}
\end{table*}

%% file: tab/table_analysis_art.tex
\begin{table}[t]
\centering
\caption{\textbf{Zero-shot performance of AI-generated image detection methods on the Art domain.} We report AUROC.}
\label{table:art}
\resizebox{0.85\textwidth}{!}{%
\begin{tabular}{cccccc}
\toprule   
Method & FatFormer \cite{liu2023fatformer} &  RIGID \cite{He24rigid} & MINDER \cite{Tsai24MINDER} & \methodname \textbf{(ours)} \\
\midrule
AUROC  & 0.531  & 0.725 & 0.365 & 0.765  \\ 
\bottomrule
\end{tabular}}
\end{table}

%% file: sec/6_conclusion.tex
\section{Conclusion}

We propose \methodname, an effective training-free AI-generated image detection method motivated by \texttt{RandomResizedCrop}, the core data augmentation scheme of self-supervised methods. \methodname shows improved performance and robustness against existing training-free methods. The main advantage of the \methodname is the ubiquity of the \texttt{RandomResizedCrop}, which enables \methodname applicable to various self-supervised models as the backbone. Since \methodname applies to vision-text trained encoders (\eg, CLIP \cite{radford2021CLIP}), our method can be extended to detecting multimodal AI-generated data. We leave this direction to future work.

\textbf{Broader Impact.} \methodname offers effective training-free detection of AI-generated images in the wild. This provides a \textit{tabula rasa} defense against the improper use of generative models, including fraud or manipulation of AI-generated images. Furthermore, our method can be applied to filter out AI-generated images from the web-scale image data.

\textbf{Limitations.} Our $\texttt{RescaleNPatchify}$ procedure induces extra computation costs due to the number of patches, although these patches can be computed in a batch-wise manner. Furthermore, \methodname is dependent on the choice of the backbone self-supervised model, and may not generalize to high-resolution real/AI-generated images outside the scope of the backbone model. Finally, when future generative models can succeed in faithfully generating realistic images (including high-frequency components) enough to fool the pre-trained foundation model, our approach can be less effective. One promising direction is to utilize recent multimodal foundation models, which may show enhanced understanding of AI-generated images. We leave this direction to future work.


%% file: sec/X_checklist.tex
\newpage
\section*{NeurIPS Paper Checklist}

\begin{enumerate}

\item {\bf Claims}
    \item[] Question: Do the main claims made in the abstract and introduction accurately reflect the paper's contributions and scope?
    \item[] Answer: \answerYes{} 
    \item[] Justification: we clearly stated the contributions and the scope in the abstract and introduction.
    \item[] Guidelines:
    \begin{itemize}
        \item The answer NA means that the abstract and introduction do not include the claims made in the paper.
        \item The abstract and/or introduction should clearly state the claims made, including the contributions made in the paper and important assumptions and limitations. A No or NA answer to this question will not be perceived well by the reviewers. 
        \item The claims made should match theoretical and experimental results, and reflect how much the results can be expected to generalize to other settings. 
        \item It is fine to include aspirational goals as motivation as long as it is clear that these goals are not attained by the paper. 
    \end{itemize}

\item {\bf Limitations}
    \item[] Question: Does the paper discuss the limitations of the work performed by the authors?
    \item[] Answer: \answerYes{} 
    \item[] Justification: we include the limitation section in the conclusion.
    \item[] Guidelines:
    \begin{itemize}
        \item The answer NA means that the paper has no limitation while the answer No means that the paper has limitations, but those are not discussed in the paper. 
        \item The authors are encouraged to create a separate "Limitations" section in their paper.
        \item The paper should point out any strong assumptions and how robust the results are to violations of these assumptions (e.g., independence assumptions, noiseless settings, model well-specification, asymptotic approximations only holding locally). The authors should reflect on how these assumptions might be violated in practice and what the implications would be.
        \item The authors should reflect on the scope of the claims made, e.g., if the approach was only tested on a few datasets or with a few runs. In general, empirical results often depend on implicit assumptions, which should be articulated.
        \item The authors should reflect on the factors that influence the performance of the approach. For example, a facial recognition algorithm may perform poorly when image resolution is low or images are taken in low lighting. Or a speech-to-text system might not be used reliably to provide closed captions for online lectures because it fails to handle technical jargon.
        \item The authors should discuss the computational efficiency of the proposed algorithms and how they scale with dataset size.
        \item If applicable, the authors should discuss possible limitations of their approach to address problems of privacy and fairness.
        \item While the authors might fear that complete honesty about limitations might be used by reviewers as grounds for rejection, a worse outcome might be that reviewers discover limitations that aren't acknowledged in the paper. The authors should use their best judgment and recognize that individual actions in favor of transparency play an important role in developing norms that preserve the integrity of the community. Reviewers will be specifically instructed to not penalize honesty concerning limitations.
    \end{itemize}

\item {\bf Theory assumptions and proofs}
    \item[] Question: For each theoretical result, does the paper provide the full set of assumptions and a complete (and correct) proof?
    \item[] Answer: \answerNA{} 
    \item[] Justification: Our paper is not based on the theoretical results but on the intuitions for AI-generated image detection. 
    \item[] Guidelines:
    \begin{itemize}
        \item The answer NA means that the paper does not include theoretical results. 
        \item All the theorems, formulas, and proofs in the paper should be numbered and cross-referenced.
        \item All assumptions should be clearly stated or referenced in the statement of any theorems.
        \item The proofs can either appear in the main paper or the supplemental material, but if they appear in the supplemental material, the authors are encouraged to provide a short proof sketch to provide intuition. 
        \item Inversely, any informal proof provided in the core of the paper should be complemented by formal proofs provided in appendix or supplemental material.
        \item Theorems and Lemmas that the proof relies upon should be properly referenced. 
    \end{itemize}

    \item {\bf Experimental result reproducibility}
    \item[] Question: Does the paper fully disclose all the information needed to reproduce the main experimental results of the paper to the extent that it affects the main claims and/or conclusions of the paper (regardless of whether the code and data are provided or not)?
    \item[] Answer: \answerYes{}{} 
    \item[] Justification: we clearly stated our hyperparameter (our algorithm is deterministic) in the experiment section. We followed the author's hyperparameters when reproducing the baseline.
    \item[] Guidelines:
    \begin{itemize}
        \item The answer NA means that the paper does not include experiments.
        \item If the paper includes experiments, a No answer to this question will not be perceived well by the reviewers: Making the paper reproducible is important, regardless of whether the code and data are provided or not.
        \item If the contribution is a dataset and/or model, the authors should describe the steps taken to make their results reproducible or verifiable. 
        \item Depending on the contribution, reproducibility can be accomplished in various ways. For example, if the contribution is a novel architecture, describing the architecture fully might suffice, or if the contribution is a specific model and empirical evaluation, it may be necessary to either make it possible for others to replicate the model with the same dataset, or provide access to the model. In general. releasing code and data is often one good way to accomplish this, but reproducibility can also be provided via detailed instructions for how to replicate the results, access to a hosted model (e.g., in the case of a large language model), releasing of a model checkpoint, or other means that are appropriate to the research performed.
        \item While NeurIPS does not require releasing code, the conference does require all submissions to provide some reasonable avenue for reproducibility, which may depend on the nature of the contribution. For example
        \begin{enumerate}
            \item If the contribution is primarily a new algorithm, the paper should make it clear how to reproduce that algorithm.
            \item If the contribution is primarily a new model architecture, the paper should describe the architecture clearly and fully.
            \item If the contribution is a new model (e.g., a large language model), then there should either be a way to access this model for reproducing the results or a way to reproduce the model (e.g., with an open-source dataset or instructions for how to construct the dataset).
            \item We recognize that reproducibility may be tricky in some cases, in which case authors are welcome to describe the particular way they provide for reproducibility. In the case of closed-source models, it may be that access to the model is limited in some way (e.g., to registered users), but it should be possible for other researchers to have some path to reproducing or verifying the results.
        \end{enumerate}
    \end{itemize}

\item {\bf Open access to data and code}
    \item[] Question: Does the paper provide open access to the data and code, with sufficient instructions to faithfully reproduce the main experimental results, as described in supplemental material?
    \item[] Answer: \answerNo{} 
    \item[] Justification: We will publish the code after the paper gets accepted. However, as our algorithm is deterministic and the benchmark is public, it would be easy to reproduce the results.
    \item[] Guidelines:
    \begin{itemize}
        \item The answer NA means that paper does not include experiments requiring code.
        \item Please see the NeurIPS code and data submission guidelines (\url{https://nips.cc/public/guides/CodeSubmissionPolicy}) for more details.
        \item While we encourage the release of code and data, we understand that this might not be possible, so “No” is an acceptable answer. Papers cannot be rejected simply for not including code, unless this is central to the contribution (e.g., for a new open-source benchmark).
        \item The instructions should contain the exact command and environment needed to run to reproduce the results. See the NeurIPS code and data submission guidelines (\url{https://nips.cc/public/guides/CodeSubmissionPolicy}) for more details.
        \item The authors should provide instructions on data access and preparation, including how to access the raw data, preprocessed data, intermediate data, and generated data, etc.
        \item The authors should provide scripts to reproduce all experimental results for the new proposed method and baselines. If only a subset of experiments are reproducible, they should state which ones are omitted from the script and why.
        \item At submission time, to preserve anonymity, the authors should release anonymized versions (if applicable).
        \item Providing as much information as possible in supplemental material (appended to the paper) is recommended, but including URLs to data and code is permitted.
    \end{itemize}

\item {\bf Experimental setting/details}
    \item[] Question: Does the paper specify all the training and test details (e.g., data splits, hyperparameters, how they were chosen, type of optimizer, etc.) necessary to understand the results?
    \item[] Answer: \answerYes{} 
    \item[] Justification: We include all the details on the hyperparameters in the experiment section.
    \item[] Guidelines:
    \begin{itemize}
        \item The answer NA means that the paper does not include experiments.
        \item The experimental setting should be presented in the core of the paper to a level of detail that is necessary to appreciate the results and make sense of them.
        \item The full details can be provided either with the code, in appendix, or as supplemental material.
    \end{itemize}

\item {\bf Experiment statistical significance}
    \item[] Question: Does the paper report error bars suitably and correctly defined or other appropriate information about the statistical significance of the experiments?
    \item[] Answer: \answerYes{} 
    \item[] Justification: Our algorithm is deterministic and does not require multiple seeds. While one baseline is based on random noise, we also experimented with the ensemble version of this.
    \item[] Guidelines:
    \begin{itemize}
        \item The answer NA means that the paper does not include experiments.
        \item The authors should answer "Yes" if the results are accompanied by error bars, confidence intervals, or statistical significance tests, at least for the experiments that support the main claims of the paper.
        \item The factors of variability that the error bars are capturing should be clearly stated (for example, train/test split, initialization, random drawing of some parameter, or overall run with given experimental conditions).
        \item The method for calculating the error bars should be explained (closed form formula, call to a library function, bootstrap, etc.)
        \item The assumptions made should be given (e.g., Normally distributed errors).
        \item It should be clear whether the error bar is the standard deviation or the standard error of the mean.
        \item It is OK to report 1-sigma error bars, but one should state it. The authors should preferably report a 2-sigma error bar than state that they have a 96\% CI, if the hypothesis of Normality of errors is not verified.
        \item For asymmetric distributions, the authors should be careful not to show in tables or figures symmetric error bars that would yield results that are out of range (e.g. negative error rates).
        \item If error bars are reported in tables or plots, The authors should explain in the text how they were calculated and reference the corresponding figures or tables in the text.
    \end{itemize}

\item {\bf Experiments compute resources}
    \item[] Question: For each experiment, does the paper provide sufficient information on the computer resources (type of compute workers, memory, time of execution) needed to reproduce the experiments?
    \item[] Answer: \answerYes{} 
    \item[] Justification: We clearly state the resource of A100GPU in the experiment section.
    \item[] Guidelines:
    \begin{itemize}
        \item The answer NA means that the paper does not include experiments.
        \item The paper should indicate the type of compute workers CPU or GPU, internal cluster, or cloud provider, including relevant memory and storage.
        \item The paper should provide the amount of compute required for each of the individual experimental runs as well as estimate the total compute. 
        \item The paper should disclose whether the full research project required more compute than the experiments reported in the paper (e.g., preliminary or failed experiments that didn't make it into the paper). 
    \end{itemize}
    
\item {\bf Code of ethics}
    \item[] Question: Does the research conducted in the paper conform, in every respect, with the NeurIPS Code of Ethics \url{https://neurips.cc/public/EthicsGuidelines}?
    \item[] Answer: \answerYes{} 
    \item[] Justification: We believe that we have faithfully practiced the code of ethics in the guideline.
    \item[] Guidelines:
    \begin{itemize}
        \item The answer NA means that the authors have not reviewed the NeurIPS Code of Ethics.
        \item If the authors answer No, they should explain the special circumstances that require a deviation from the Code of Ethics.
        \item The authors should make sure to preserve anonymity (e.g., if there is a special consideration due to laws or regulations in their jurisdiction).
    \end{itemize}

\item {\bf Broader impacts}
    \item[] Question: Does the paper discuss both potential positive societal impacts and negative societal impacts of the work performed?
    \item[] Answer: \answerYes{} 
    \item[] Justification: We discuss the broader impact in the conclusion section.
    \item[] Guidelines:
    \begin{itemize}
        \item The answer NA means that there is no societal impact of the work performed.
        \item If the authors answer NA or No, they should explain why their work has no societal impact or why the paper does not address societal impact.
        \item Examples of negative societal impacts include potential malicious or unintended uses (e.g., disinformation, generating fake profiles, surveillance), fairness considerations (e.g., deployment of technologies that could make decisions that unfairly impact specific groups), privacy considerations, and security considerations.
        \item The conference expects that many papers will be foundational research and not tied to particular applications, let alone deployments. However, if there is a direct path to any negative applications, the authors should point it out. For example, it is legitimate to point out that an improvement in the quality of generative models could be used to generate deepfakes for disinformation. On the other hand, it is not needed to point out that a generic algorithm for optimizing neural networks could enable people to train models that generate Deepfakes faster.
        \item The authors should consider possible harms that could arise when the technology is being used as intended and functioning correctly, harms that could arise when the technology is being used as intended but gives incorrect results, and harms following from (intentional or unintentional) misuse of the technology.
        \item If there are negative societal impacts, the authors could also discuss possible mitigation strategies (e.g., gated release of models, providing defenses in addition to attacks, mechanisms for monitoring misuse, mechanisms to monitor how a system learns from feedback over time, improving the efficiency and accessibility of ML).
    \end{itemize}
    
\item {\bf Safeguards}
    \item[] Question: Does the paper describe safeguards that have been put in place for responsible release of data or models that have a high risk for misuse (e.g., pretrained language models, image generators, or scraped datasets)?
    \item[] Answer: \answerNA{} 
    \item[] Justification: We do not release any data or model that pose high risk for misuse.
    \item[] Guidelines:
    \begin{itemize}
        \item The answer NA means that the paper poses no such risks.
        \item Released models that have a high risk for misuse or dual-use should be released with necessary safeguards to allow for controlled use of the model, for example by requiring that users adhere to usage guidelines or restrictions to access the model or implementing safety filters. 
        \item Datasets that have been scraped from the Internet could pose safety risks. The authors should describe how they avoided releasing unsafe images.
        \item We recognize that providing effective safeguards is challenging, and many papers do not require this, but we encourage authors to take this into account and make a best faith effort.
    \end{itemize}

\item {\bf Licenses for existing assets}
    \item[] Question: Are the creators or original owners of assets (e.g., code, data, models), used in the paper, properly credited and are the license and terms of use explicitly mentioned and properly respected?
    \item[] Answer: \answerYes{} 
    \item[] Justification: We cite the respective benchmarks for AI-generated image detection. Furthermore, we also include the methods that create the dataset and the original datasets. License of these datasets are provided in the Appendix.
    \item[] Guidelines:
    \begin{itemize}
        \item The answer NA means that the paper does not use existing assets.
        \item The authors should cite the original paper that produced the code package or dataset.
        \item The authors should state which version of the asset is used and, if possible, include a URL.
        \item The name of the license (e.g., CC-BY 4.0) should be included for each asset.
        \item For scraped data from a particular source (e.g., website), the copyright and terms of service of that source should be provided.
        \item If assets are released, the license, copyright information, and terms of use in the package should be provided. For popular datasets, \url{paperswithcode.com/datasets} has curated licenses for some datasets. Their licensing guide can help determine the license of a dataset.
        \item For existing datasets that are re-packaged, both the original license and the license of the derived asset (if it has changed) should be provided.
        \item If this information is not available online, the authors are encouraged to reach out to the asset's creators.
    \end{itemize}

\item {\bf New assets}
    \item[] Question: Are new assets introduced in the paper well documented and is the documentation provided alongside the assets?
    \item[] Answer: \answerNA{} 
    \item[] Justification: The paper does not release new assets. We will publish our code if accepted.
    \item[] Guidelines:
    \begin{itemize}
        \item The answer NA means that the paper does not release new assets.
        \item Researchers should communicate the details of the dataset/code/model as part of their submissions via structured templates. This includes details about training, license, limitations, etc. 
        \item The paper should discuss whether and how consent was obtained from people whose asset is used.
        \item At submission time, remember to anonymize your assets (if applicable). You can either create an anonymized URL or include an anonymized zip file.
    \end{itemize}

\item {\bf Crowdsourcing and research with human subjects}
    \item[] Question: For crowdsourcing experiments and research with human subjects, does the paper include the full text of instructions given to participants and screenshots, if applicable, as well as details about compensation (if any)? 
    \item[] Answer: \answerNA{} 
    \item[] Justification: We do not involve any crowdsourcing or research with human subjects.
    \item[] Guidelines:
    \begin{itemize}
        \item The answer NA means that the paper does not involve crowdsourcing nor research with human subjects.
        \item Including this information in the supplemental material is fine, but if the main contribution of the paper involves human subjects, then as much detail as possible should be included in the main paper. 
        \item According to the NeurIPS Code of Ethics, workers involved in data collection, curation, or other labor should be paid at least the minimum wage in the country of the data collector. 
    \end{itemize}

\item {\bf Institutional review board (IRB) approvals or equivalent for research with human subjects}
    \item[] Question: Does the paper describe potential risks incurred by study participants, whether such risks were disclosed to the subjects, and whether Institutional Review Board (IRB) approvals (or an equivalent approval/review based on the requirements of your country or institution) were obtained?
    \item[] Answer: \answerNA{} 
    \item[] Justification: We do not report any result that involves human subject.
    \item[] Guidelines:
    \begin{itemize}
        \item The answer NA means that the paper does not involve crowdsourcing nor research with human subjects.
        \item Depending on the country in which research is conducted, IRB approval (or equivalent) may be required for any human subjects research. If you obtained IRB approval, you should clearly state this in the paper. 
        \item We recognize that the procedures for this may vary significantly between institutions and locations, and we expect authors to adhere to the NeurIPS Code of Ethics and the guidelines for their institution. 
        \item For initial submissions, do not include any information that would break anonymity (if applicable), such as the institution conducting the review.
    \end{itemize}

\item {\bf Declaration of LLM usage}
    \item[] Question: Does the paper describe the usage of LLMs if it is an important, original, or non-standard component of the core methods in this research? Note that if the LLM is used only for writing, editing, or formatting purposes and does not impact the core methodology, scientific rigorousness, or originality of the research, declaration is not required.
    \item[] Answer: \answerNA{} 
    \item[] Justification: Our methodology is not based on any LLMs.
    \item[] Guidelines:
    \begin{itemize}
        \item The answer NA means that the core method development in this research does not involve LLMs as any important, original, or non-standard components.
        \item Please refer to our LLM policy (\url{https://neurips.cc/Conferences/2025/LLM}) for what should or should not be described.
    \end{itemize}

\end{enumerate}

%% file: sec/Y_Appendix.tex
\input{algorithm/pseudocode}

\section{Appendix}

\subsection{Pseudocode of WarPAD}

We show the Pytorch-like pseudocode of \methodname in Algorithm \ref{alg:code}. Note that all operations allow batch-wise computation, hence we can process the input patches in a batch-wise manner.

\subsection{Further information of the Experiment settings.}

\textbf{Synthbuster.} The Synthbuster benchmark consists of 1000 real RAISE-1k images and 9000 AI-generated images consisting of scene and art images under the 'CC BY-NC-SA 4.0' license. We download all real \footnote{\url{https://loki.disi.unitn.it/RAISE/download.html}} and AI-generated datasets \footnote{\url{https://zenodo.org/records/10066460}} in the URL via the author's official repository.

\textbf{GenImage.} The GenImage benchmark consists of ImageNet real data and AI-generated data consisting of 8 different generative models under the 'CC BY-NC-SA 4.0' license. Each test consists of pairs of real and AI-generated image pairs, where the size is 6000+6000 except of SDv1.5, where the size is 8000+8000. We download the datasets via the author's official repository \footnote{\url{https://github.com/GenImage-Dataset/GenImage}}.

\textbf{Deepfake-LSUN-Bedroom.} The Deepfake-LSUN-Bedroom benchmark consists of 10000 real LSUN-Bedroom images and 10 $\times$ 10000 AI-generated data where the model is trained to generate LSUN-Bedroom-like images. We download the datasets via the author's official repository \footnote{\url{https://zenodo.org/records/7528113}} under the 'CC BY 4.0' license.

\textbf{Baselines.} We follow the author's implementation for the AEROBLADE \footnote{\url{https://github.com/jonasricker/aeroblade}} and Manifold Bias \footnote{\url{https://tinyurl.com/zeroshotimplementation}}, respectively. Since the original implementation of the AEROBLADE operates on the fixed dimension, extension to data with arbitrary size (\eg, Synthbuster, GenImage) is not trivial. Our finding is that the preservation of the original dimension is crucial for the performance, hence we chose to center-crop or resize the image to the fixed dimension of the autoencoder dimension whether the image is larger or smaller than the autoencoder default dimension, respectively. On the other hand, we have not found any official implementation of the authors on the RIGID and MINDER. Instead, we manage to reproduce the RIGID in the consistent setting of our \methodname. Note that RIGID and MINDER propose to resize the image to the default resolution of DINOv2, which is 224 $\times$ 224.

\textbf{Experiment Settings.} Most experiments are deterministic since they operate on deterministic operations. A slight exception is RIGID, where the Gaussian noise augmentation is done on the image. However, our experiments in Table~\ref{table:ablate} show that RIGID with multiple runs does not change much performance against RIGID with a single seed.

\textbf{Additional Backbones.} All pre-trained backbones are accessible and downloadable. For the CLIP model, we use "clip-vit-base-32" \footnote{\url{https://huggingface.co/openai/clip-vit-base-patch32}} for the base model. We use SwaV on the Resnet50 \cite{he2016resnet} backbone \footnote{\url{https://github.com/facebookresearch/swav}} and DINO of "vit-s16" version \footnote{\url{https://github.com/facebookresearch/dino}} pre-trained on the ImageNet dataset. We use the "vit-mae-base" model for the ViTMAE backbone \footnote{\url{https://huggingface.co/docs/transformers/en/model_doc/vit_mae}} and "beit-base-patch16-224" model for the BeiT backbone \footnote{\url{https://huggingface.co/docs/transformers/en/model_doc/beit}}.

\input{fig/fig_wavelet_choice}

\subsection{Histogram of other wavelet choices.}

We show the computed histogram of WaRPAD on other wavelets (db2, coif1, bior3.1, and coif3), on real and SDv1.4-generated images computed in the Synthbuster benchmark in Figure~\ref{fig:appendix_wavelet}. Results show DINOv2 model loses its robustness in other wavelet choices, especially wavelets with more vanishing moments.

\input{fig/fig_robustness_appendix}

\subsection{Further robustness experiments}

We further include the performance of \methodname, AEROBLADE, MINDER, and RIGID in the Synthbuster benchmark in Figure~\ref{fig:robustness_synthbuster} consistent to Figure~\ref{fig:robustness_genimage}. The trend is consistent, where the \methodname performs the best.

%% file: algorithm/pseudocode.tex
\makeatletter
\lst@Key{spacestyle}{}
  {\def\lst@visiblespace{{#1\lst@ttfamily{\char32}\textvisiblespace{}}}}
\makeatother

\lstset{
  backgroundcolor=\color{white},
  basicstyle=\fontsize{9pt}{9pt}\ttfamily\selectfont,
  columns=fullflexible,
  breaklines=true,
  captionpos=b,
  commentstyle=\fontsize{9pt}{9pt}\color{codeblue},
  keywordstyle=\fontsize{9pt}{9pt}\color{codekw},
  showspaces=true,
  showstringspaces=false,
  spacestyle   = \color{white},
}

\begin{algorithm}[ht]
\caption{\methodname (PyTorch-like Pseudo-code)}\label{algorithm:pseudocode}
\label{alg:code}
\definecolor{codeblue}{rgb}{0.25,0.5,0.5}
\definecolor{codekw}{rgb}{0.85, 0.18, 0.50}
\begin{lstlisting}[language=python]
# f(x): normalized [cls] token output of self-supervised model
# alpha: weight of perturbation
# DWTForward, DWTInverse: forward and inverse discrete wavelet transform
# Sim: cosine similarity function

def HFwav(x):
    x_low, x_high = DWTForward(x)
    N_perturb = DWTInverse([torch.zeros_like(x_low), x_high])
    feat_original = f(x)
    feat_perturb = f(x - alpha * N_perturb)
    return Sim(feat_original, feat_perturb)

def WaRPAD(x):
    x_patch = RescaleNPatchify(x)
    f_patch = HFwav(x_patch)
    return f_patch.mean()
    
\end{lstlisting}
\end{algorithm}


%% file: fig/fig_wavelet_choice.tex
\begin{figure}[t]
    \centering
    \begin{subfigure}[b]{0.19\textwidth}
        \includegraphics[width=\textwidth]{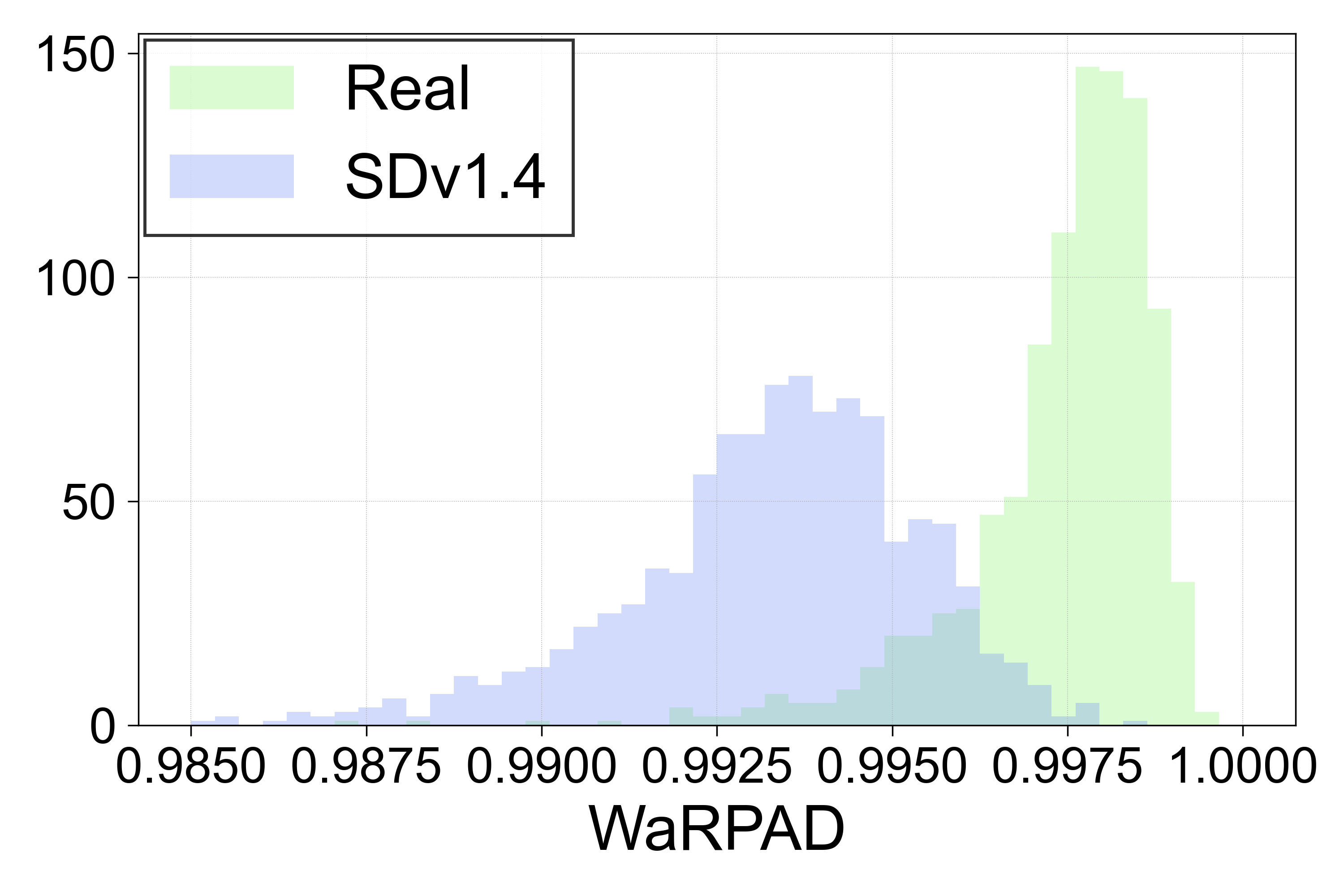}
        \caption{Haar}
        \label{subfigure:appendix_haar}
    \end{subfigure}
    \hfill
    \begin{subfigure}[b]{0.19\textwidth}
        \includegraphics[width=\textwidth]{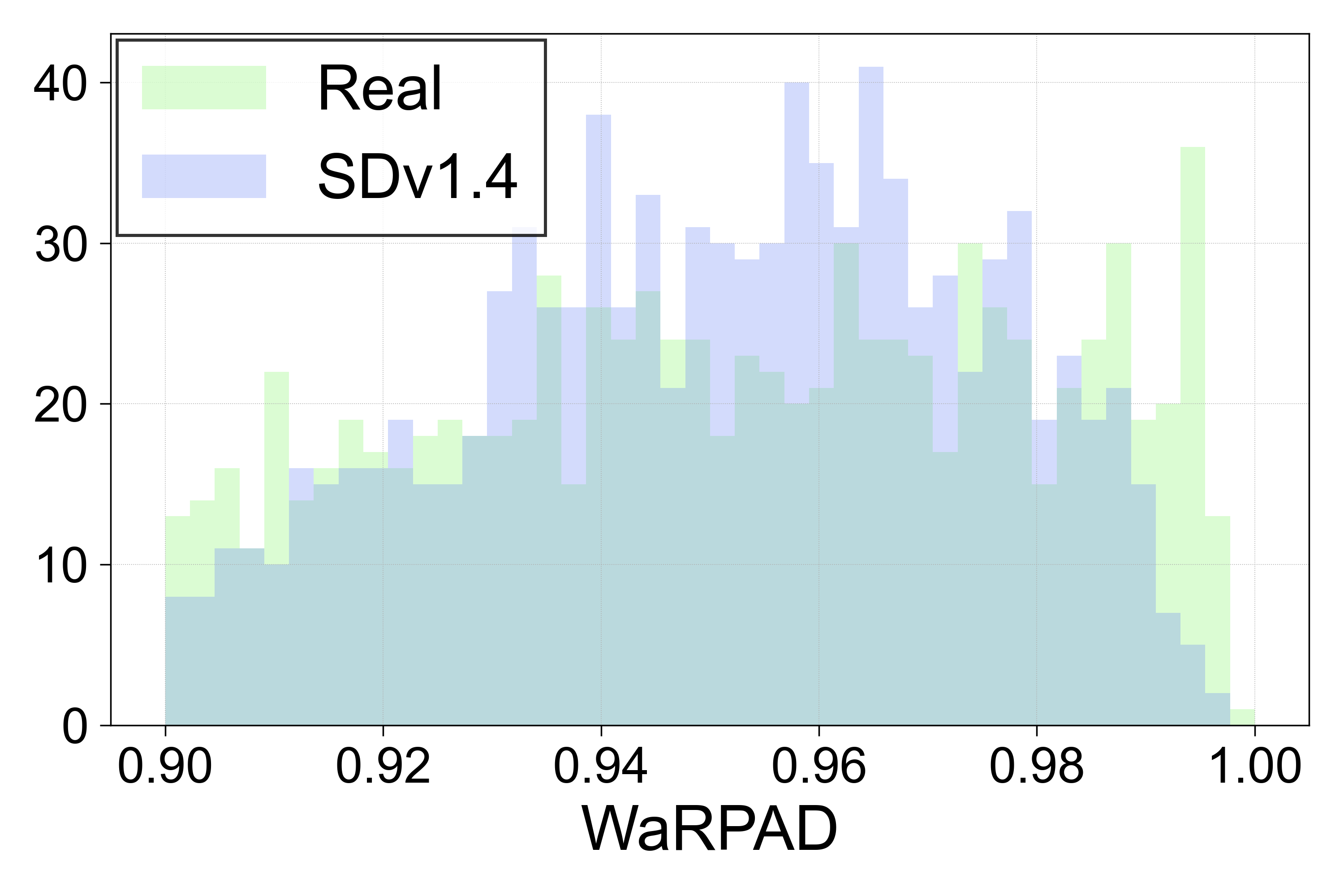}
        \caption{db2}
        \label{subfigure:appendix_db2}
    \end{subfigure}
    \hfill    
    \begin{subfigure}[b]{0.19\textwidth}
        \includegraphics[width=\textwidth]{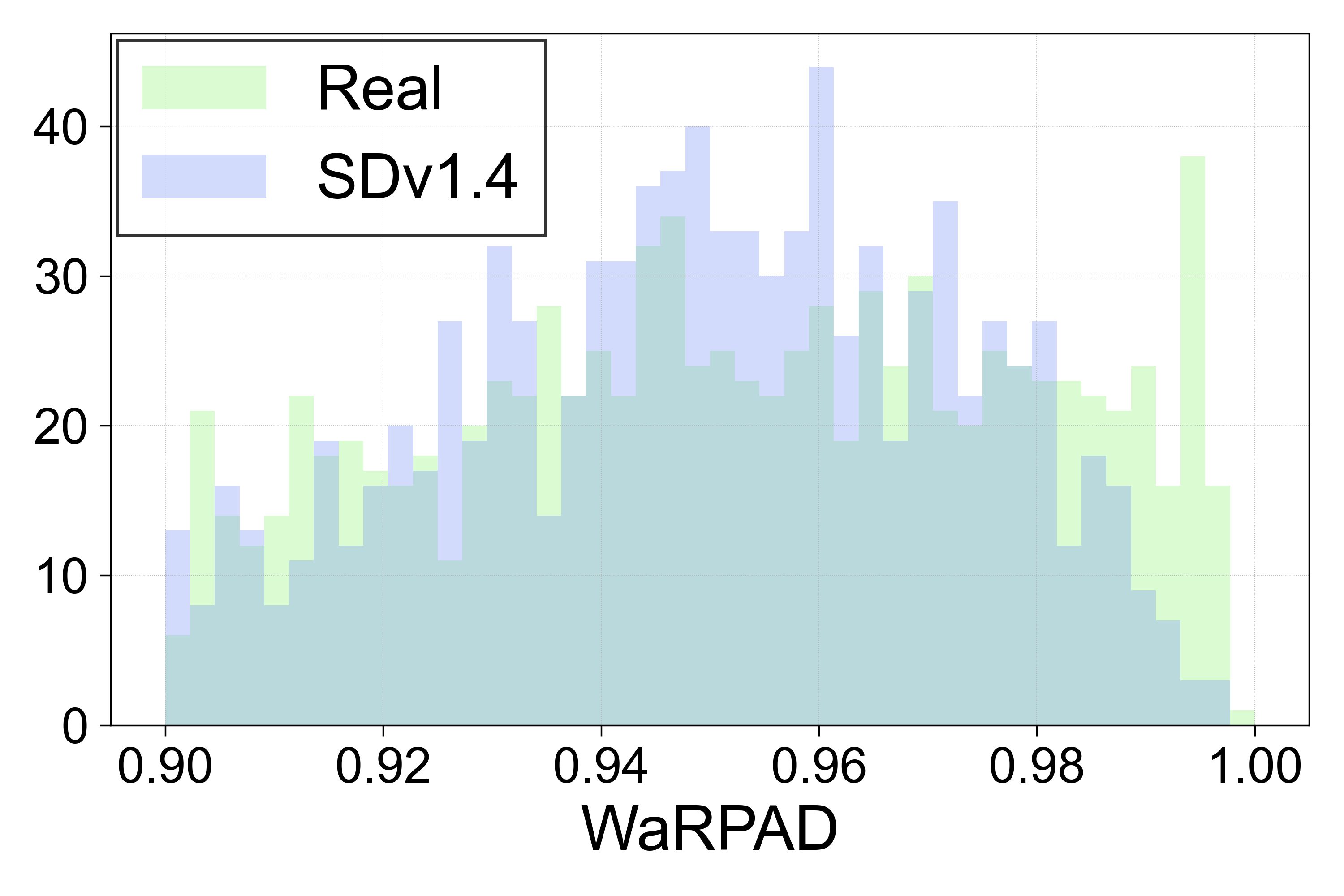}
        \caption{coif1}
        \label{subfigure:appendix_coif1}
    \end{subfigure}
    \begin{subfigure}[b]{0.19\textwidth}
        \includegraphics[width=\textwidth]{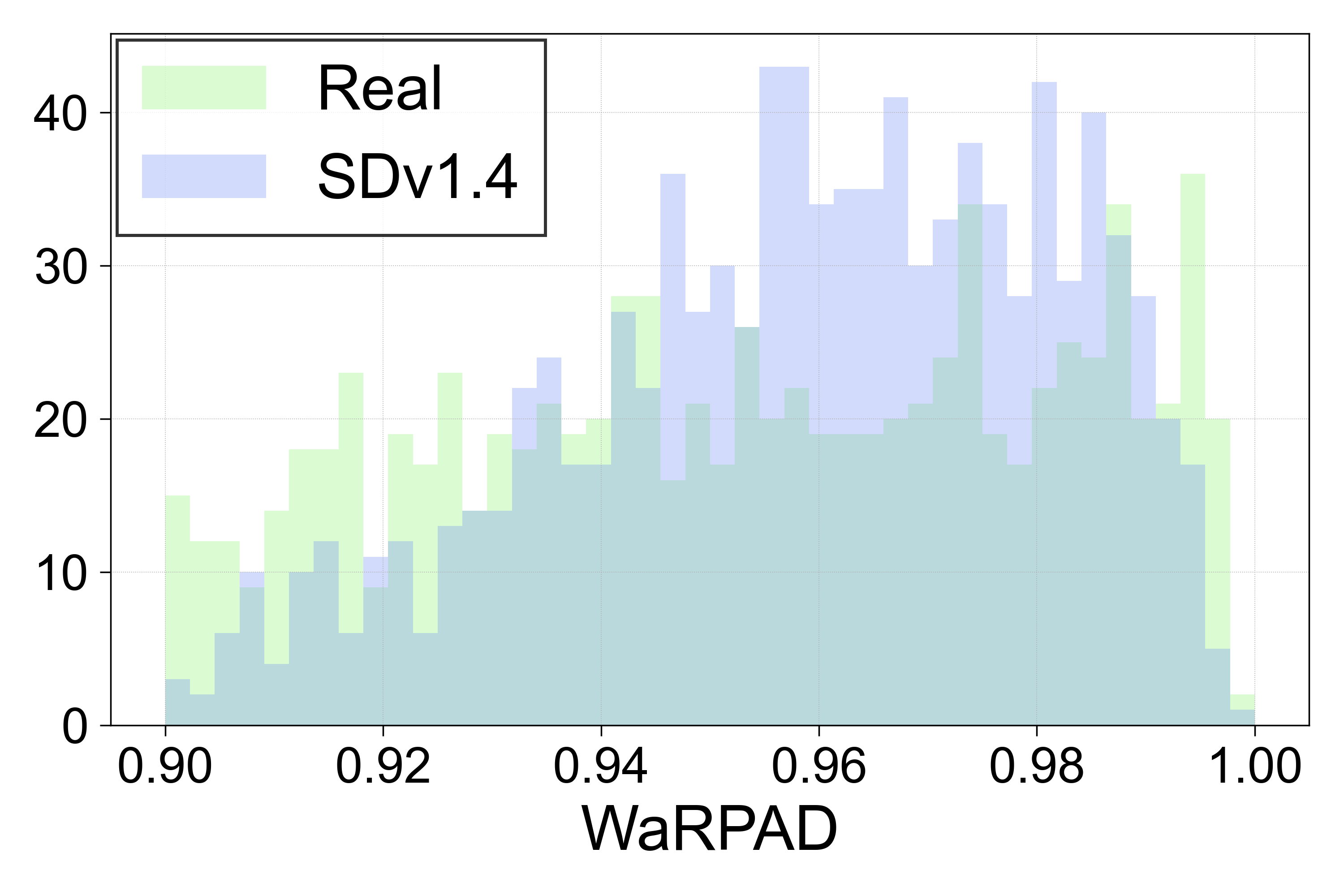}
        \caption{bior3.1}
        \label{subfigure:appendix_bor3.1}
    \end{subfigure}
    \hfill    
    \begin{subfigure}[b]{0.19\textwidth}
        \includegraphics[width=\textwidth]{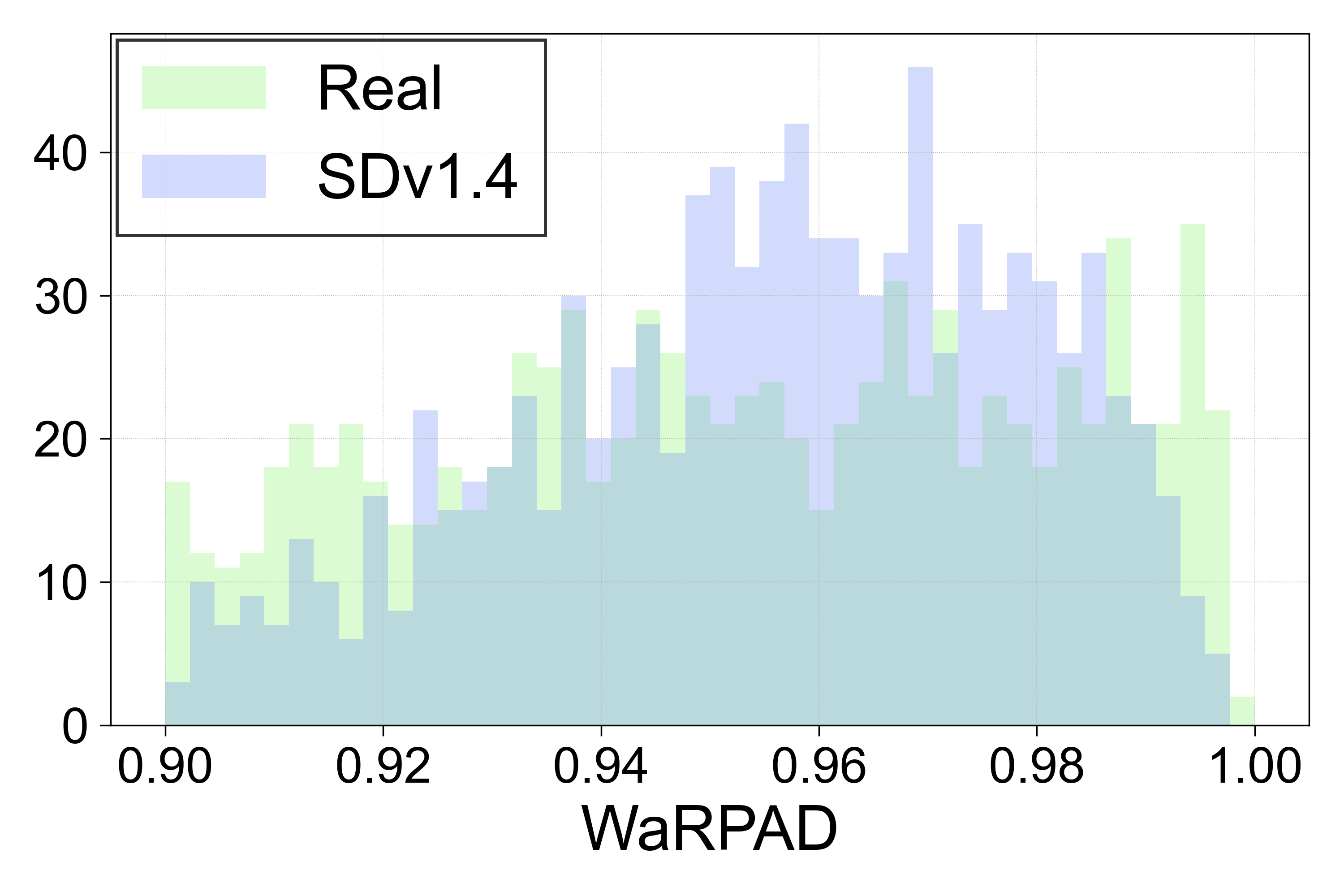}
        \caption{coif3}
        \label{subfigure:appendix_coif3}
    \end{subfigure}
    \caption{\textbf{Histogram of other wavelet.} We show the results on Haar, db2, coif1,  bior3.1, and coif3 wavelet, respectively.} 
    \label{fig:appendix_wavelet}
\vspace{-0.1in}
\end{figure}

%% file: fig/fig_robustness_appendix.tex
\begin{figure}[t]
    \centering
    \begin{subfigure}[b]{0.32\textwidth}
        \includegraphics[width=\textwidth]{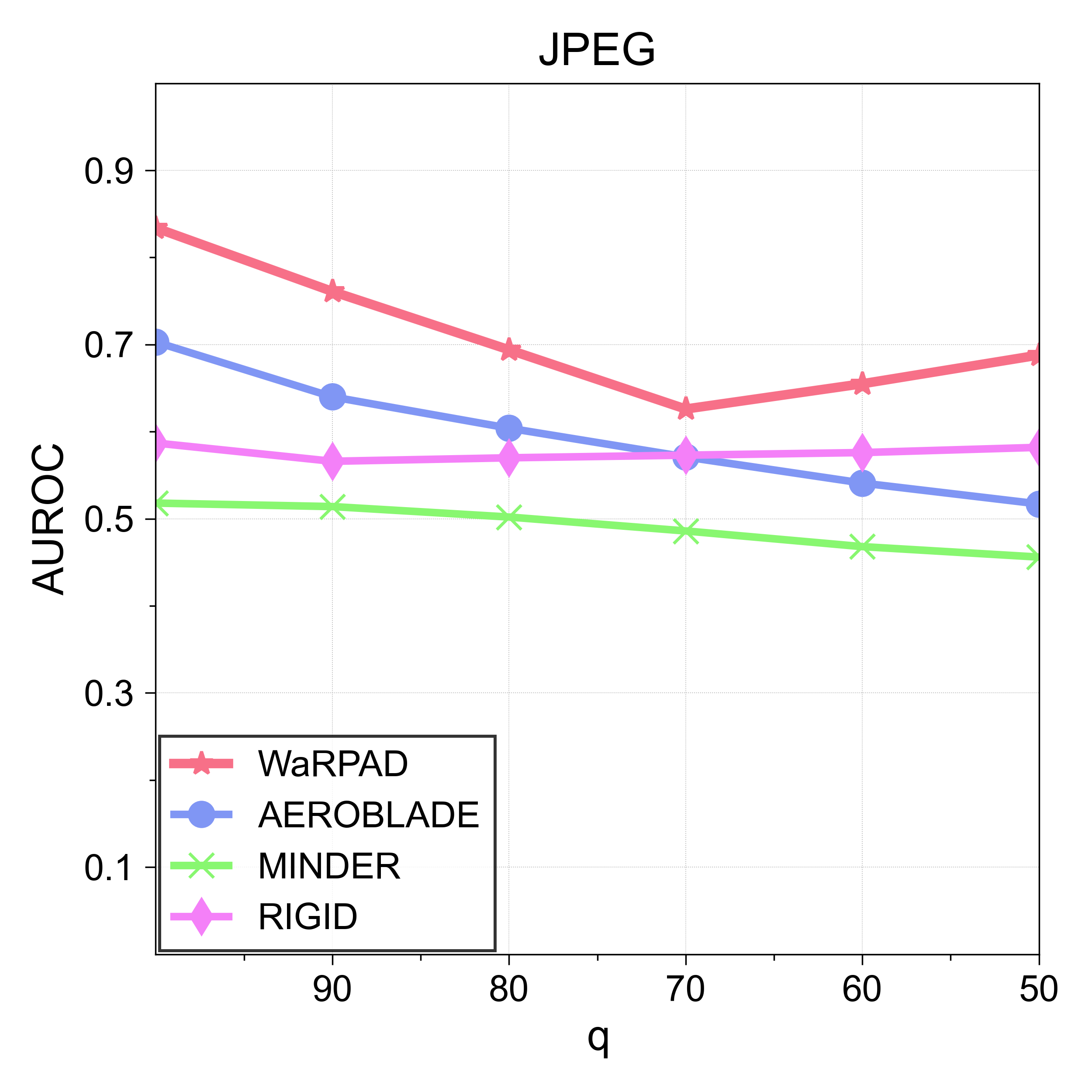}
        \caption{JPEG Compression}
    \end{subfigure}
    \hfill
    \begin{subfigure}[b]{0.32\textwidth}
        \includegraphics[width=\textwidth]{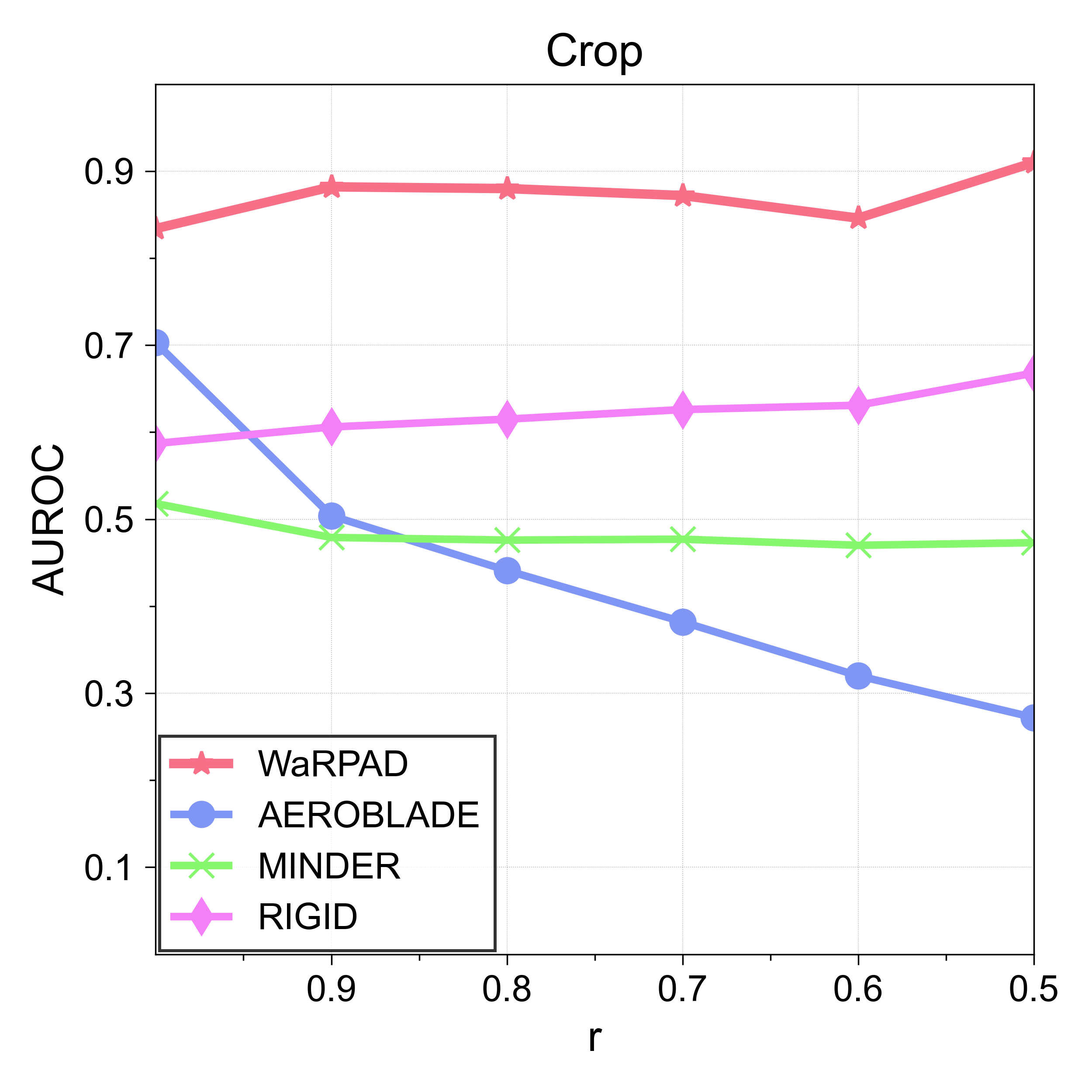}
        \caption{Center Crop}
    \end{subfigure}
    \hfill     
    \begin{subfigure}[b]{0.32\textwidth}
        \includegraphics[width=\textwidth]{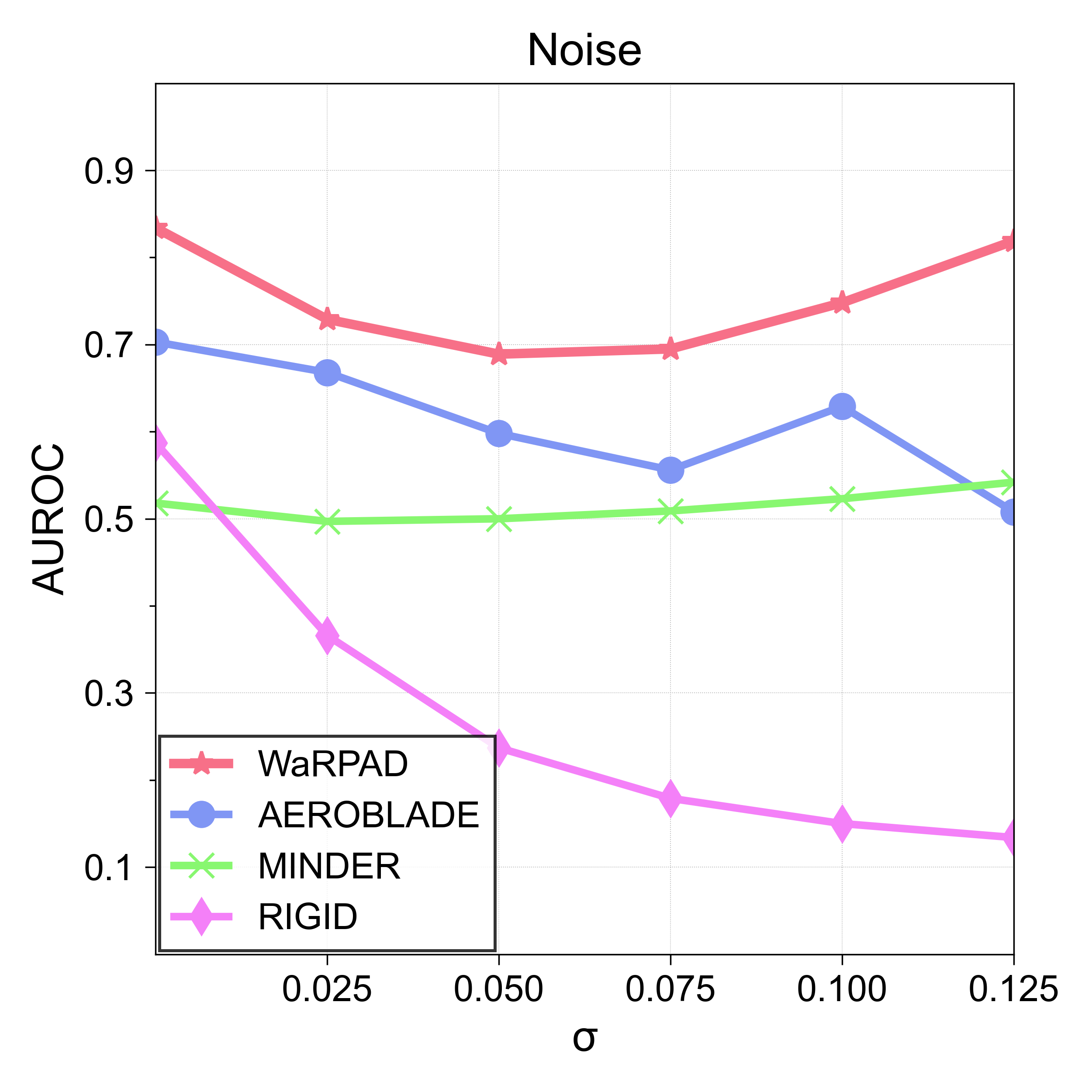}
        \caption{Gaussian Noise}
    \end{subfigure}
    \caption{\textbf{Robustness of \methodname in corruptions.} We test the AUROC performance of \methodname, AEROBLADE, MINDER, and RIGID in corrupted test images in the Synthbuster benchmark.}
    \label{fig:robustness_synthbuster}
\vspace{-0.1in}
\end{figure}